\documentclass[twoside,11pt]{article}

\usepackage{blindtext}

\usepackage{jmlr2e}

\usepackage[utf8]{inputenc} 
\usepackage[T1]{fontenc}    
\usepackage{hyperref}       
\usepackage{url}            
\usepackage{booktabs}       
\usepackage{amsfonts}       
\usepackage{nicefrac}       
\usepackage{microtype}      
\usepackage{xcolor}         

\usepackage{amsmath,amsfonts,bm}

\def\eqref#1{eq.~(\ref{#1})}

\def\1{\bm{1}}

\DeclareMathAlphabet{\mathsfit}{\encodingdefault}{\sfdefault}{m}{sl}
\SetMathAlphabet{\mathsfit}{bold}{\encodingdefault}{\sfdefault}{bx}{n}

\newcommand{\R}{\mathbb{R}}

\DeclareMathOperator*{\argmax}{arg\,max}
\DeclareMathOperator*{\argmin}{arg\,min}

\usepackage{amssymb,bm}

\usepackage{url}
\usepackage{comment}

\usepackage[utf8]{inputenc} 
\usepackage[T1]{fontenc}    
\usepackage{url}            
\usepackage{amsfonts}       
\usepackage{nicefrac}       
\usepackage{xcolor}         
\usepackage{arydshln}
\usepackage{pdfpages}

\usepackage{adjustbox}
\usepackage{caption}
\usepackage{subcaption}
\usepackage{listings}

\usepackage{tikz}
\usetikzlibrary{arrows,positioning,chains,shapes.geometric,fit,backgrounds}

\tikzstyle{inputNode}=[draw,circle,minimum size=10pt,inner sep=0pt]
\tikzstyle{stateTransition}=[-stealth, thick]

\tikzset{decision/.style={diamond, draw, fill=blue!20, text width=2.5em, text badly centered, inner sep=0pt}}
\tikzset{block/.style={rectangle, draw, fill=blue!20, text width=5em, text centered, rounded corners,
 minimum width=0.25cm}}
 \tikzset{block/.style={rectangle, draw, fill=blue!20, text width=5em, text centered, rounded corners,
 minimum width=0.25cm}}
 \tikzset{block/.style={rectangle, draw, fill=blue!20, text width=5em, text centered, rounded corners,
 minimum width=0.25cm}}
\tikzset{line/.style={draw, -latex}}

\usepackage{colortbl} 
\usepackage{multirow}

\usepackage{mathtools,xparse}
\usepackage{nicefrac}       
\usepackage{wrapfig}

\usepackage{algpseudocode, algorithm}
\makeatletter
\renewcommand{\ALG@beginalgorithmic}{\small}
\makeatother

\usepackage{cleveref}

\newcommand{\qedsymbol}{\hfill$\square$}

\usepackage{thmtools, thm-restate}

\usepackage{lastpage}
\jmlrheading{23}{2022}{1-\pageref{LastPage}}{1/21; Revised 5/22}{9/22}{21-0000}{Holtz, Chen, Cloninger, Cheng, and Mishne}

\ShortHeadings{Semi-Supervised Laplace Learning on Stiefel Manifolds}{Holtz, Chen, Cloninger, Cheng, and Mishne}
\firstpageno{1}

\begin{document}

\title{Semi-Supervised Laplace Learning on Stiefel Manifolds}

\author{\name Chester Holtz \email chholtz@ucsd.edu \\
  \addr Department of Computer Science\\
  University of California San Diego\\
  La Jolla, CA \\
  \AND
  Pengwen Chen \email pengwen@email.nchu.edu.tw \\
  \addr Department of Applied Mathematics \\
  National Chung Hsing University \\
  South District, Taichung, Taiwan
  \AND 
  Alexander Cloninger \email acloninger@ucsd.edu \\
  \addr Department of Mathematics\\
  University of California San Diego\\
  La Jolla, CA  
  \AND 
  Chung-Kuan Cheng \email ckcheng@ucsd.edu \\
  \addr Department of Computer Science\\
  University of California San Diego\\
  La Jolla, CA 
  \AND 
  Gal Mishne \email gmishne@ucsd.edu \\
  \addr Halicio\v{g}lu Data Science Institute\\ 
  University of California San Diego \\
  La Jolla, CA
}

\editor{My editor}

\maketitle

\begin{abstract}
Motivated by the need to address the degeneracy of canonical Laplace learning algorithms in low label rates, we propose to reformulate graph-based semi-supervised learning as a nonconvex generalization of a \emph{Trust-Region Subproblem} (TRS).
This reformulation is motivated by the well-posedness of Laplacian eigenvectors in the limit of infinite unlabeled data.
To solve this problem, we first show that a first-order condition implies the solution of a manifold alignment problem and that solutions to the classical \emph{Orthogonal Procrustes} problem can be used to efficiently find good classifiers that are amenable to further refinement. To tackle refinement, we develop the framework of Sequential Subspace Optimization for graph-based SSL. Next, we address the criticality of selecting supervised samples at low-label rates. We characterize informative samples with a novel measure of centrality derived from the principal eigenvectors of a certain submatrix of the graph Laplacian. We demonstrate that our framework achieves lower classification error compared to recent state-of-the-art and classical semi-supervised learning methods at extremely low, medium, and high label rates.
\end{abstract}

\begin{keywords}
  semi-supervised learning, graph learning, active learning, optimization
\end{keywords}

\section{Introduction}

Semi-supervised methods leverage both labeled and unlabeled data for tasks such as classification and regression.
In semi-supervised learning (SSL), we are given a partially-labeled training set consisting of both labeled examples and unlabeled examples. The goal is to leverage the unlabeled examples to learn a predictor that is better than a predictor that is trained using the labeled examples alone. This setup is motivated by the high cost of obtaining annotated data in practical problems. Consequently, we are typically interested in the regime where the number of labeled examples is significantly smaller than the number of training points.
For problems where very few labels are available, the geometry of the unlabeled data can be used to significantly improve the performance of classic machine learning models.
Additionally, the choice of labeled vertices is also a critical factor in this regime. In this work, we introduce a unified framework for graph-based semi-supervised and active learning at low label rates.

An important work in graph-based semi-supervised learning is Laplace learning~\citep{zhu2003semi}, which seeks a harmonic function that extends provided labels over the unlabeled vertices. Laplace learning, and its variants (notably, Poisson Learning~\citep{calder20poisson}) have been widely applied in semi-supervised and graph-structured learning~\citep{zhou2005learning,zhou2004learning,ando2007learning,yang2006sslmanifold}.

In this work, we improve upon the state-of-the-art for graph-based semi-supervised learning at very low label rates. Classical Laplace learning and label propagation algorithms yield poor classification results~\citep{nadler2009semi,el2016asymptotic} in this regime. This is typically attributed to the fact that the solutions develop localized spikes near the labeled vertices and are nearly constant for vertices distant from labels. In other words, Laplace learning-based algorithms often fail to adequately propagate labels over the graph, given few labeled nodes. To address this issue, recent work has suggested imposing small adjustments to classical Laplace learning procedure. For example, $p$-Laplace learning~\citep{el2016asymptotic,slepcev2019analysis,calder2018game,calder2017consistency} for $p>2$, and particularly for $p=\infty$, often yields superior empirical performance compared to Laplace learning at low label rates~\citep{flores2019algorithms}. 
Other relevant methods for addressing low label rate problems include higher-order Laplacian regularization \citep{zhou2011semi} and spectral classification~\citep{Belkin2002UsingMS, zhou2011error}.


In addition to our classifier, we describe a simple active-learning strategy that exploits certain computational elements of our algorithm. The majority of existing active learning strategies typically involve evaluating the informativeness of unlabeled samples. For example, one of the most commonly used query frameworks is uncertainty sampling \citep{burractivelearning2012, miller2022graphbasedal, miller2021modelchange, jivopt2012} where the active learner queries the data samples that it is most uncertain about how to label. Most general uncertainty sampling strategies use some notion of margin as a measure of uncertainty \citep{burractivelearning2012, miller2022graphbasedal}.

Many active learning algorithms that excel at low-label rates also employ strategies based on the connectivity of the graph, e.g., the degree centrality or cut structure \citep{bianchial2013,guillorylabelcut2009,ma2023partitionbased}. Related work includes geometric landmarking methods, which seek to maximize coverage of the collected samples. For example, \citep{maxgeodist,ajinkya2018distancesampling} propose geodesic distance-based strategies to greedily add new landmarks with large cumulative geodesic distance to existing landmarks. However, these methods are computationally prohibitive on most benchmarks. Particularly relevant to our work are algebraic landmarking methods. In particular, \cite{Xu2015ActiveML} proposed an algebraic reconstruction error bound based on the Gershgorin circle theorem (GCT)~\citep{gershgorin} and an associated greedy algorithm based on this bound. However, this method suffers from high complexity due to logarithmic computations of a large matrix. 

\subsection{Contribution}

In this work, we propose to solve a natural semi-supervised extension of Laplacian Eigenmaps and spectral cuts, which are well-posed in the limit of unlabeled data. Our extension is motivated by an optimization-based perspective of Laplacian Eigenmaps as a Rayleigh Quotient minimization problem over all labeled and unlabeled vertices. We show that a natural partitioning of the problem yields a more general quadratically constrained quadratic program over the unlabeled vertices. We then generalize the sequential subspace (SSM) framework originally proposed to solve similar problems in $\mathbb{R}^n$ to $\mathbb{R}^{n\times k}$ and we develop an associated active learning scheme.

To summarize, our contributions are:
\begin{enumerate}
    \item 
    We introduce a natural formulation of graph semi-supervised learning as a rescaled quadratic program on a compact Stiefel Manifold, i.e. a generalization of a \emph{Trust-Region Subproblem}. 
    \item We describe a scalable approximate method, globally convergent iterative methods, and a graph cut-based refinement scheme to solve this problem and demonstrate robustness in a variety of label rate regimes.
    \item We introduce a score to characterize informative samples based on the principal eigenvectors of the \emph{grounded Laplacian} and relate this score to a particular absorbing random walk defined on the graph. An estimate of the score is obtained ``for free'' from the SSM subproblem. 
    \item We compare our approach to competing semi-supervised graph learning algorithms and demonstrate state-of-the-art performance in low, medium, and high label rate settings on MNIST, Fashion-MNIST, and CIFAR-10. 
\end{enumerate}
The rest of the paper is organized as follows.  In Section \ref{sec:preliminaries} we briefly introduce Laplacian Eigenmaps and our supervised variant, and then provide a detailed motivation for the algorithm. Our formulation is presented in Section \ref{sec:formulation}. Approximate and iterative algorithms are presented in Section \ref{sec:algorithms} and our approach to active learning at low label rates is presented in Section \ref{sec:graphal}. In Section \ref{sec:experiments} we present numerical experiments. We conclude and discuss future work in Section \ref{sec:conclusion}.
\section{Preliminaries and notations}
\label{sec:preliminaries}
We assume the data can be viewed as lying on a graph, such that each vertex is a data-point. 
Let $\mathcal{V} = \{v_1, v_2,\ldots , v_M\}$ denote the $M$ vertices of the graph with edge weights $w_{ij} \geq 0$ between $v_i$ and $v_j$. We assume that the graph is symmetric, so $w_{ij} = w_{ji}$. The degree of a vertex is defined as $d_i = \sum_{j=1}^n w_{ij}$ . 

For a multi-class classification problem with $k$ classes, we let the standard basis vector $e_i \in \mathbb{R}^k$ represent the $i$-th class (i.e. a ``one-hot encoding''). Without loss of generality, we assume the first $m$ vertices $l = \{v_1, v_2,\ldots , v_m \}$ are given labels $y_1, y_2,\ldots , y_m \in \{e_1, e_2,\ldots , e_k\}$, where $m \ll M$. Let $n$ denote the number of unlabeled vertices, i.e. $n = M-m$. The problem of graph-based semi-supervised learning is to smoothly propagate the labels over the unlabeled vertices $\mathcal{U} = \{v_{m+1}, v_{m+2},\ldots , v_M\}$. 
The compact \emph{Stiefel Manifold} is denoted
\begin{equation}\label{eq:stiefel}
\textrm{St}(n,k) = \{X\in \mathbb{R}^{n\times k} : X^\top X = I\}.
\end{equation}
Note that the projection of a matrix $X \in \mathbb{R}^{n\times k}$ onto $\textrm{St}(n,k)$, denoted $[X]_+ := \argmin\{||X_s - X||_F : X_s \in \textrm{St}(n,k)\}$ is given by 
\begin{equation}
\label{rem:stiefel_projection}
    [X]_+ = UV^\top,
\end{equation}
where $X = U\Sigma V^\top$ is the rank-$k$ truncated singular value decomposition of $X$. 
Given a graph and a set of labeled vertices, the Laplace learning algorithm \citep{zhu2003semi} extends the labels over the graph by solving the following problem
\begin{equation}\label{eq:bv}
\left.\begin{aligned}
x(v_i) &=y_i,&&\text{if }1 \leq i \leq m \\
(\mathcal{L} x)_i &= 0,&&\text{if }m+1 \leq i \leq M
\end{aligned}\right\}
\end{equation}
where $\mathcal{L}$ is the unnormalized graph Laplacian given by $\mathcal{L}=D-W$, $D$ is a diagonal matrix whose elements are the node degrees,
and $x:\mathcal{V}\to \R^k$. 
The prediction for vertex $v_i$ is determined by the largest component of $x(v_i)$:
\begin{equation}\label{eq:labeldec}
\argmax_{j\in \{1,\dots,k\}} \{x_j(v_i)\}.
\end{equation}
Note that Laplace learning is also called \emph{label propagation (LP)} \citep{zhu2005semi}, since the Laplace equation \eqref{eq:bv}, can be solved by repeatedly replacing $x(v_i)$ with the weighted average of its neighbors.

The solution of Laplace learning is the minimizer of the following problem with label constraints $x(v_i) = y_i$:
\begin{equation}\label{eq:laplace}
\min_{x \in \mathbb{R}^M} \left\{ x^\top \mathcal{L}x : x(v_i) = y_i,\:\: 1\leq i \leq m \right\}
\end{equation}

We assume $k$ is a positive integer (much) less than $n$. Let $K$ denote the set $\{1, 2,\ldots , k\}$. Let $I_{n,k}$ denote the submatrix of the identity matrix $I_n$, consisting of the first $k$ columns. Let $O_k$ be the orthogonal group, i.e., $Q \in O_k$ if and only if $Q \in \mathbb{R}_{k,k}$ and $Q^\top Q = I_k$.  Let $\langle A, B \rangle$ be the trace of the matrix $A^\top B$. Let $\mathbf{1}$ denote the all-ones vector.
\subsection{Spectral Embeddings with Supervision}
\label{sec:formulation}
In Laplacian Eigenmaps~\citep{belkin09laplacianeigenmaps}, one seeks an embedding of the graph vertices via the eigenfunctions of the graph Laplacian corresponding to the smallest nontrivial eigenvalues. Equivalently, this can be expressed as the following \emph{Quadratically Constrained Quadratic Program} (QCQP) over the vertices of the graph:
\begin{equation}
\min_{X_0}\langle X_0, \mathcal{L} X_0 \rangle \quad
\text{s.t. } X_0^\top X_0 = I,\:\: \mathbf{1}^\top X_0 = 0.
\label{eq:lap_eig}
\end{equation}
The notation $X_0 \in \mathbb{R}^{M\times k}$ is the mapping of the $M$ vertices to a $k$-dimensional space. 
In the case where $k=1$, \eqref{eq:lap_eig} is also known in the numerical analysis literature as a \emph{Rayleigh quotient minimiziation problem}~\citep{GoluVanl96}. Despite its nonconvexity, a unique (up to orthogonal transformations) global solution is given by the set of eigenvectors of $\mathcal{L}$ corresponding to the smallest $k$ nontrivial (nonzero) eigenvalues of $\mathcal{L}$.

We first extend this framework with supervision, similarly to Laplace learning in~\eqref{eq:laplace}. Additionally, to facilitate the supervised decomposition, we rescale $I$ uniformly by $p=M / k$, the balanced proportion of samples associated with each class:
\begin{equation}\label{eq:supervised_lapeig}
\begin{aligned}
\min_{X_0}\langle X_0, \mathcal{L} X_0 \rangle \quad 
\text{s.t. } X_0^\top X_0 = pI,\:\: &\mathbf{1}^\top X_0 = 0, \:\:(X_0)_i = y_i,\:\: i\in[m]
\end{aligned}
\end{equation}
The associated prediction is then $\ell(x_i) = \argmax_{j\in \{1,\dots,k\}} (X_0)_{ij}$. 
Next, we show how supervision naturally leads to a partitioning of the problem. We denote the submatrices of $X_0$ and $\mathcal{L}$ corresponding to the $n$ unlabeled vertices $\mathcal{U} \subseteq \mathcal{V}$ and $m$ labeled vertices $l \subseteq \mathcal{V}$ as $X_\mathcal{U}$, $X_l$ and $\mathcal{L}_\mathcal{U}$, $\mathcal{L}_l$, respectively. More concretely, $
\mathcal{L} = \left[\begin{smallmatrix}
\mathcal{L}_{l} & \mathcal{L}_{l\mathcal{U}}\\
\mathcal{L}_{\mathcal{U}l} & \mathcal{L}_{\mathcal{U}}
\end{smallmatrix}\right]
$ and 
$X_0 = \left[\begin{smallmatrix}
X_{l}\\
X_{\mathcal{U}}
\end{smallmatrix}\right]$. 

In general, addressing the quadratic and linear equality constraints pose a significant challenge from an optimization standpoint. We propose to address this by solving an equivalent rescaled problem. As demonstrated in the proposition below, via careful substitution to eliminate the linear constraint, we show how the problem may be rescaled and efficiently and robustly solved as a quadratic program on a compact \emph{Stiefel Manifold}.
The associated solution to this problem can then be used to determine the labels of the unlabeled vertices, as in Laplace learning (\eqref{eq:bv}).

\begin{proposition}
Let $p$ be a positive scalar. Consider the minimization
\begin{equation}\label{eq:cneqi}
    \min_{X_{\mathcal{U}}\in\mathbb{R}^{n\times k}} \left \{\langle X_0, \mathcal{L} X_0 \rangle : X_0 = [X_l^\top; X_\mathcal{U}^\top]^\top, X_0^\top X_0 = pI, \mathbf{1}^\top X_0 = 0  \right\}
\end{equation}
Let $r = -X_l^\top \mathbf{1}$ and 
$P = I - \frac{1}{n}\mathbf{1}\mathbf{1}^\top$ and 
\begin{equation}
    L = P \mathcal{L}_{\mathcal{U}}P,\:\: B = P(\mathcal{L}_{\mathcal{U}l}X_l + \frac{1}{n} \mathcal{L}_\mathcal{U}\mathbf{1}r^\top),\:\: C = pI - X_l^\top X_l - \frac{1}{n} rr^\top.
\end{equation}
Then, $X_\mathcal{U} = XC^{1/2} + \frac{1}{n} \mathbf{1} r^\top$, where $X$ is the minimizer of 
\begin{equation}
    \min_{X\in\mathbb{R}^{n\times k}}  \left \{\langle X, L X C \rangle - 2\langle X, B C^{1/2} \rangle : X \in St(n,k) \right\}
\end{equation}
\end{proposition}

\noindent \emph{Proof. }  To eliminate the linear constraint, we introduce two substitutions: first, let $(X_\mathcal{U}')_i = (X_\mathcal{U})_i -   \frac{1}{n}r^\top$ denote a row-wise centering transformation with respect to the labeled nodes. This implies $\mathbf{1}^\top X_\mathcal{U}' = 0$ and also implies the quadratic constraint $X^\top X = C := pI - X_l^\top X_l - \frac{1}{n}rr^\top$ for $X$. More concretely, the first moment condition $\mathbf{1}^\top X_0 = 0$ yields 
\begin{equation}
    X_\mathcal{U}^\top \mathbf{1} = -X_l^\top \mathbf{1} =: r
\end{equation}
Second, we introduce the projection $P= I - \frac{1}{n}\mathbf{1}\mathbf{1}^\top$ onto the subspace orthogonal to the vector $\mathbf{1}\in \mathbb{R}^{n}$, i.e., $\mathbf{1}^\top(PX_\mathcal{U}') = 0$, which maps iterates onto the set of matrices with mean-zero columns.
To obtain a solution limited to this subspace, we introduce the substitutions $B= P (\mathcal{L}_{\mathcal{U}l}X_l + \mathcal{L}_\mathcal{U} \frac{1}{n}\mathbf{1}r^\top)$ and $L = P\mathcal{L}_\mathcal{U}P$ which implies $\mathbf{1}^\top B = 0$. Thus, $X_\mathcal{U}'$ is the solution of
\begin{equation}
    \min_{X_\mathcal{U} \in \mathbb{R}^{n\times k}} \left\{ \langle X_\mathcal{U}' , L X_\mathcal{U}' \rangle + 2 \langle X_\mathcal{U}', B \rangle + \text{constant} \right\}
\end{equation}
subject to
\begin{equation}
    {X_\mathcal{U}'}^\top X'_\mathcal{U} = C := p I - X_l^\top X_l - \frac{1}{n} rr^\top \in \mathbb{R}^{k\times k}
\end{equation}
The proof is completed by considering the substitution $X_\mathcal{U}' = X C^{1/2}$. \hfill \qedsymbol

By the above proposition, graph-based semi-supervised learning is equivalent to the following rescaled problem.
\begin{equation}\label{eq:rescaled_f}
\begin{aligned}
\min_{X :X\in \textrm{St}(n,k)} \left\{ F(X) = \langle X, L X C \rangle - 2\langle X, B C^{1/2} \rangle \right\}.
\end{aligned}
\end{equation}
To reiterate, given a solution to \eqref{eq:rescaled_f}, $X^*$, one recovers a solution to \eqref{eq:cneqi} via the transformation $X^* C^{1/2} + \frac{1}{n}\mathbf{1}r^\top$. This is the key formulation of this paper. 
%
%

Note that \eqref{eq:rescaled_f} is a generalization of well-known problems that arise in trust-region methods, optimization of a nonconvex quadratic over a unit ball or sphere~\citep{sorensen82newtontrust,conn00trust}, i.e. problems of the form
$$
\min_{x\in\mathbb{R}^n :||x|| = 1} \langle x, Lx \rangle - \langle x, b \rangle .
$$
We define the Lagrangian of \eqref{eq:rescaled_f} where $\Lambda\in \mathbb{R}^{k\times k}$ are the Lagrange multipliers:
\begin{equation}\label{eq:rescaled_f_lagrangian}
\begin{aligned}
\langle X, LXC \rangle - \langle X, BC^{1/2} \rangle 
- \langle \Lambda, (X^\top X - I) \rangle .
\end{aligned}
\end{equation}
The first-order condition is then
\begin{equation}\label{eq:rescaled_f_foc}
\begin{aligned}
LXC = BC^{1/2} + X\Lambda 
\end{aligned}
\end{equation}
for some $\Lambda$. Solutions $X$ that satisfy \eqref{eq:rescaled_f_foc} are \emph{critical points} or \emph{stationary points}. In general, there could exist many critical points that satisfy this condition. In general, at these ``stationary points'' (maximizers, minimizers, or saddle points), (1.) the eigenvalues of $\Lambda$ characterize the optimality of $X$ and (2.) finding good critical points necessitates computation of the eigenvectors of $L$.
\section{Semi-Supervised Spectral Learning Algorithms}
\label{sec:algorithms}

In this section, we introduce approximate and iterative methods to solve \eqref{eq:rescaled_f}. 
In theory, one can start with an arbitrary initialization to obtain a critical point of \eqref{eq:rescaled_f} using a variety of projection- or retraction-based gradient methods, with the descent direction given by the gradient of \eqref{eq:rescaled_f} and the polar projection onto the Stiefel manifold given by \eqref{rem:stiefel_projection}. However, the empirical rate of convergence depends significantly on the initialization of the embedding matrix $X$.
In order to improve convergence of our method, we first introduce and motivate an efficient method based on Procrustes Analysis~\citep{wang08manifold} to approximately compute critical points of the \emph{unscaled} objective ($C = I$).
This approximation is appropriate in the limit of few labeled examples or unlimited unlabeled data
: 
since $C = (p-\widetilde{p})I - \frac{\widetilde{p}^2}{n}\mathbf{1}\mathbf{1}^\top$, where $\widetilde{p} = m/k$, then $C \approx pI$. For example, on MNIST with one labeled vertex per class, $1/p \cdot C$ consists of a diagonal term with entries $0.9998$ and off-diagonal terms with entries $-2.778 \times 10^{-9}$. Likewise, when the number of labeled vertices per class is increased to $100$, The diagonal term reduces to $0.998$, and the off-diagonal term reduces to order $10^{-7}$. 

\subsection{Efficient approximation via Orthogonal Procrustes}
\label{sec:approx}
Here we propose an efficient way to compute approximate critical points of \eqref{eq:rescaled_f}. As previously mentioned, quadratic optimization over the Stiefel manifold is a nonconvex problem. Finding good initializations is necessary for fast convergence of iterative methods. First we solve the canonical eigenvalue problem $\min_X \text{tr}(X^\top LX)$ subject to a constraint on the second moment of $X$: $X^\top X = I$, yields $X$ are the eigenvectors of $L$.
Second, we appropriately transform the solution so that $X^\top B$ is positive definite (i.e. satisfies a necessary condition for first-order optimality). 
\begin{proposition}[Definiteness conditions of $X^\top B$]
Assume $C = I$. 
Note the first term of the objective in \eqref{eq:rescaled_f} satisfies the invariance $\langle X, LX\rangle = \langle \widetilde{X},  L\widetilde{X}\rangle$, where $\widetilde{X} = XQ$ for any orthogonal $Q \in \mathbb{R}^{k\times k}$. 
Suppose $\widetilde{X}$ is a local minimizer of \eqref{eq:rescaled_f}. Then, $\widetilde{X}^\top B\succcurlyeq 0$ and symmetric. 
\label{rem:negdef}
\end{proposition}
\emph{Proof.} By assumption, $X$ and $\widetilde{X}$ are feasible\textemdash i.e. $\widetilde{X}^\top \widetilde{X} = X^\top X = I$.
Since $C = I$, $$F(\widetilde{X}) = F(XQ) = \langle XQ, LXQ \rangle - \langle Q, X^\top B \rangle.$$
Fix $X$. 
Note that the first term satisfies the invariance $\langle X, LX \rangle = \langle XQ, LXQ \rangle = \langle \widetilde{X}, L\widetilde{X} \rangle$. For any orthogonal $Q \in \mathbb{R}^{k\times k}$. The optimal choice of $Q$ is determined by the second term. A standard result from matrix analysis yields its minimizer~\citep{horn13}. Let $X^\top B = UDV^\top$ be the SVD of $X^\top B$. Then, $Q = U_B V_B^\top$ and $\langle \widetilde{X}, B\rangle = \langle XQ, B\rangle = \langle Q, X^\top B \rangle = \langle I, D \rangle = tr(D) \geq 0$.

Therefore, $\widetilde{X}^\top B = (XQ)^\top B = Q^\top X^\top B = VU^\top UDV^\top = VDV^\top$ is symmetric and positive definite. \hfill\qedsymbol

A consequence of this is the following: Let the SVD of $X^\top B = U_BD_BV_B^\top$ and let $Q = U_BV_B^\top$. 
Algorithmically, this implies that projecting $X$ onto $Q$ decreases the objective of \eqref{eq:rescaled_f} (assuming $C=I$). Note that in practice we can additionally rescale predictions by taking $X\gets XC^{1/2}$ to properly observe the constraint on the second moment of $X$.  

\begin{figure}[t]
\centering
\begin{subfigure}[b]{0.21\linewidth}
\resizebox{4cm}{3cm}{\scalebox{1.0}[1.0]{\reflectbox{\includegraphics[width=1.0\linewidth]{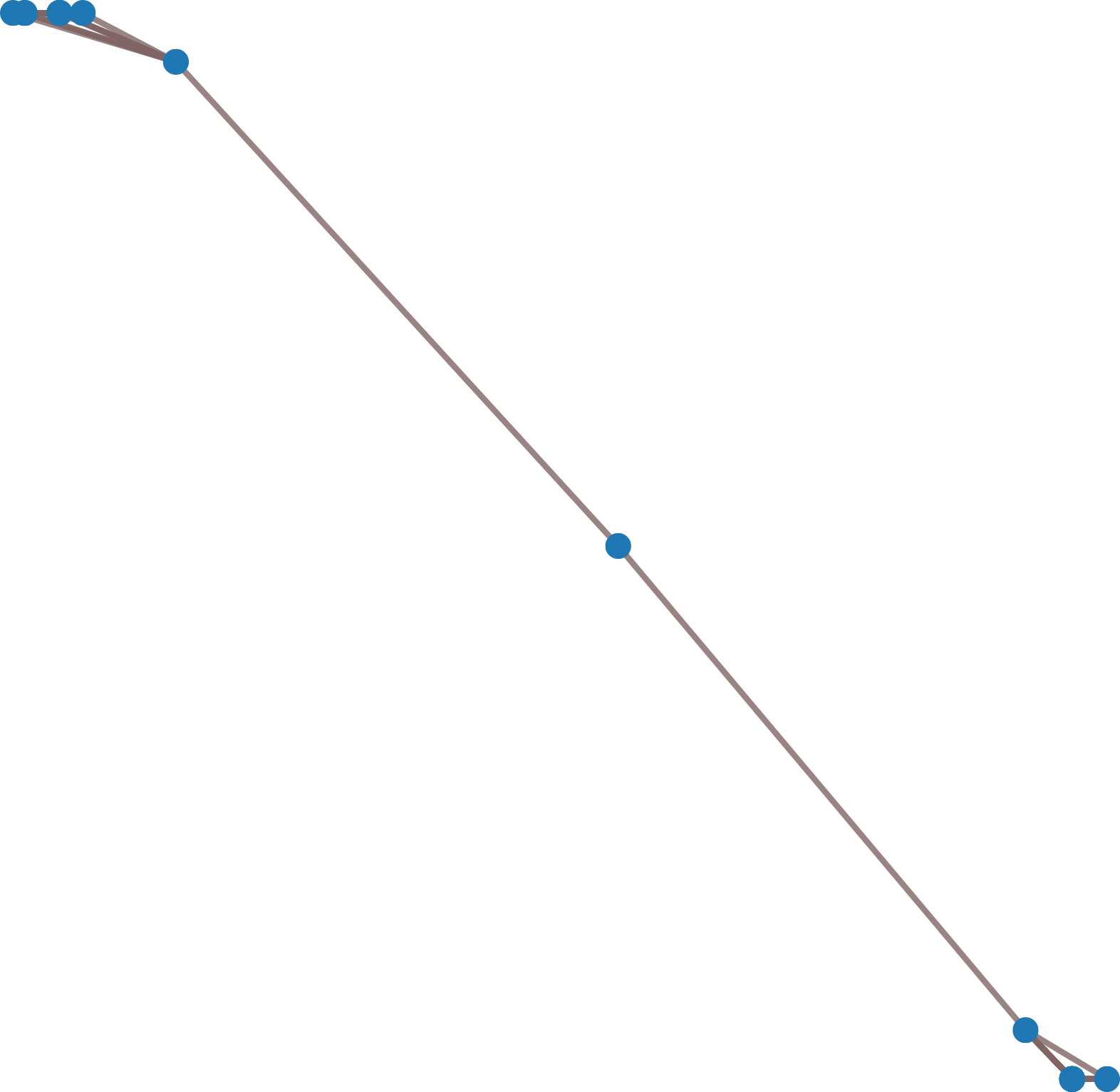}}}}
\subcaption{}
\label{fig:grid1}
\end{subfigure}
\begin{subfigure}[b]{0.2\linewidth}
\resizebox{4cm}{3cm}{\scalebox{1.0}[1.062]{\includegraphics[width=1.0\linewidth]{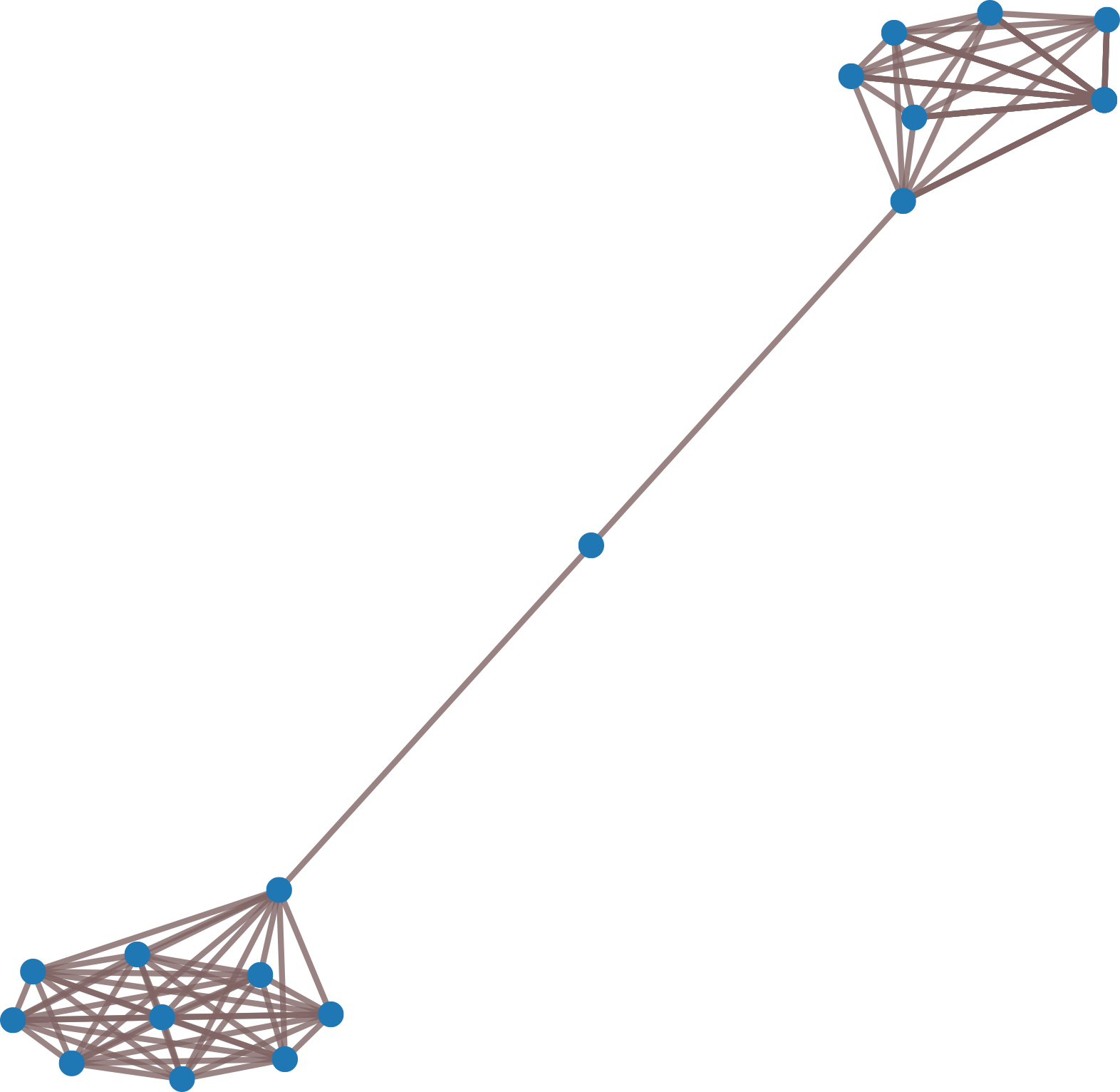}}}
\subcaption{}
\label{fig:grid2}
\end{subfigure}
%
%
\begin{subfigure}[b]{0.21\linewidth}
\resizebox{4cm}{3cm}{\scalebox{1.0}[1.0]{\includegraphics[width=1.0\linewidth]{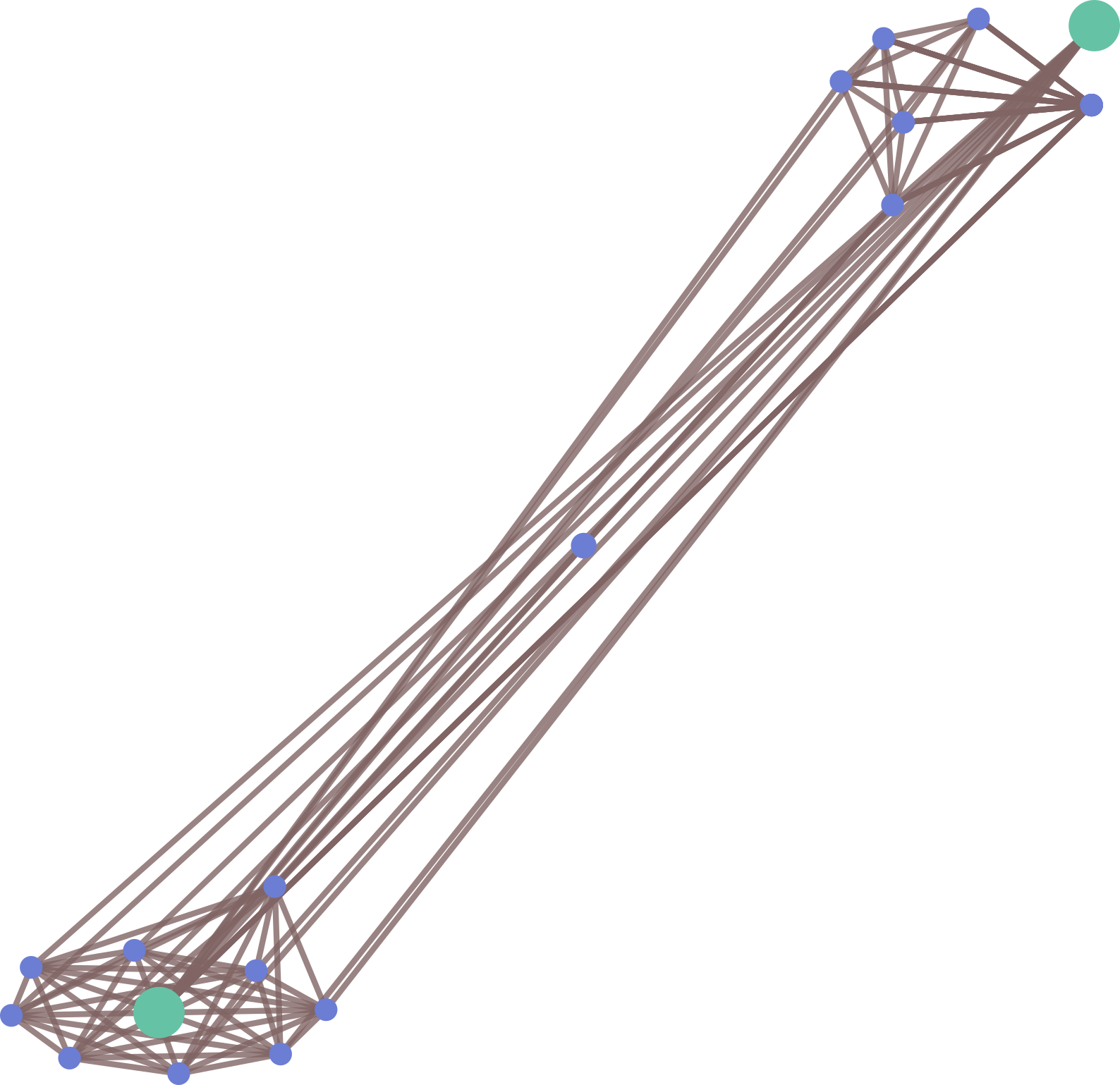}}}
\subcaption{}
\label{fig:grid3}
\end{subfigure}
\begin{subfigure}[b]{0.2\linewidth}
\resizebox{4cm}{3cm}{\scalebox{1.0}[1.062]{\includegraphics[width=1.0\linewidth]{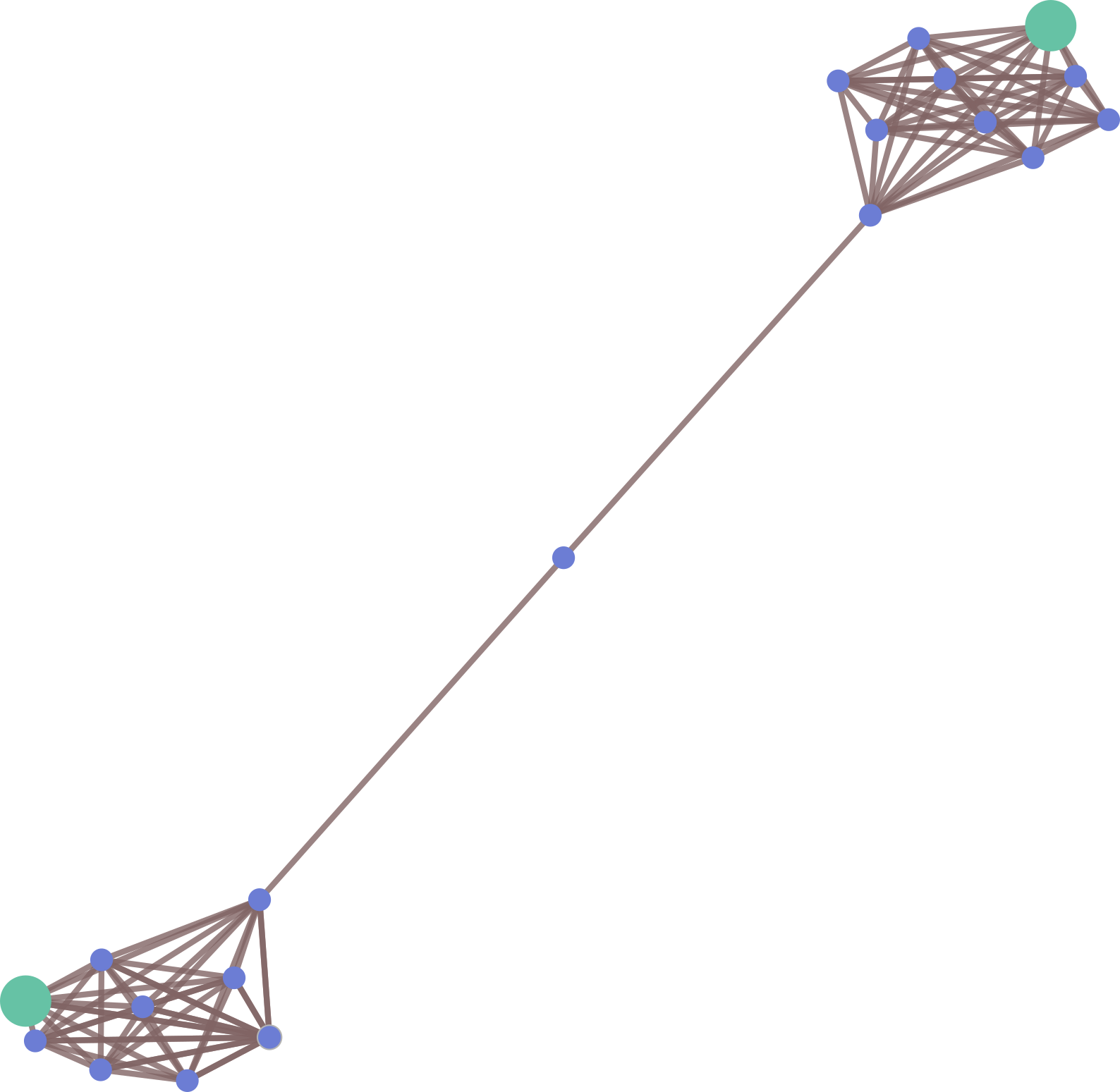}}}
\subcaption{}
\label{fig:grid4}
\end{subfigure}
%
\caption{\textbf{Eigenvector method and projection example on the barbell graph}. (\textbf{\subref{fig:grid1}}): Embedding into $\mathbb{R}^k$ via Laplacian Eigenmaps. (\textbf{\subref{fig:grid2}}): Several iterations of gradient-based repulsion are applied to remove vertex overlaps for better visualization.
(\textbf{\subref{fig:grid3}}): Consider taking an arbitrary vertex from each clique and assigning it a label (green vertices). Spectral embeddings are likely \emph{inconsistent} with labeled vertices. (\textbf{\subref{fig:grid4}}): Procrustes embedding. The orthogonal transform $Q$ is derived from Prop. \ref{rem:negdef} and applied to $X$; $XQ$ resolves the discrepancy between the embeddings and the labeled vertices.}
\label{fig:procrustes}
\vspace{-0.5cm}
\end{figure}

This projection step can be interpreted as an alignment step, where we find an orthogonal transformation $Q$ that aligns unlabeled vertices with their neighboring labeled vertices. This transformation is then applied to all unlabeled vertices.  
We briefly describe the connection with Orthogonal Procrustes Analysis~\citep{wang08manifold}. Let $X$ be feasible, i.e. $X^\top X = I$. Note that the invariance is nothing but $\text{tr}(X^\top L X) = \langle X, LX \rangle = \langle XQ, LXQ \rangle$ for any orthogonal $Q$. Thus, 
%
\begin{equation}\label{eq:procrustes}
\begin{aligned}
\argmin_{Q :Q\in \textrm{St}(k,k)}\langle XQ, LXQ \rangle - \langle XQ, B \rangle 
=\argmax_{Q :Q\in \textrm{St}(k,k)} \langle XQ, B \rangle 
=\argmin_{Q :Q\in \textrm{St}(k,k)} ||XQ - B||^2_F. 
\end{aligned}
\end{equation}
This problem is the canonical Orthogonal Procrustes problem in $\mathbb{R}^{k\times k}$ in the context of finding an alignment between the axis-aligned labeled vertices and their neighborhood of unlabeled vertices. We demonstrate the effect of this procedure in Figure~\ref{fig:procrustes}. In Figure~\ref{fig:procrustes}(\subref{fig:grid1}),(\subref{fig:grid2}), we plot the first pair of eigenvectors corresponding to the smallest two nonzero eigenvalues associated with a barbell graph. In Figure~\ref{fig:procrustes}(\subref{fig:grid3}), we pick a random pair of vertices $v_i$ and $v_j$ with coordinates $x_i$ and $x_j$ from each clique and assign labels $y_i = x_i$ and $y_j = x_j$. Under this labeling, we say that the embedding is \emph{inconsistent}. 
We then show that by applying the approximate method based on Procrustes Analysis introduced in Section \ref{sec:approx}, we recover an embedding which is \emph{consistent} with the labels.

Alternatively, the projection and $Q$-transform
%
%
can be interpreted as 
performing orthogonal multivariate regression in the space spanned by the first $k$ nontrivial eigenvectors of $L$: 
\begin{equation}\label{eq:orthoreg}
Q = \argmin_{Q :Q\in \textrm{St}(k,k)}\sum_{i\in [m]}||x_iQ - y_i||_2^2, 
\end{equation}
where $Q \in \mathbb{R}^{k\times k}$ and predictions are given by $XQ$. Note that this is similar in principle to the Semi-Supervised Laplacian Eigenmaps (SSL) algorithm~\citep{Belkin2002UsingMS}, which solves an ordinary least squares problem using eigenvectors $X$ of the Laplacian as features:
$$
Q = \argmin_{Q}\sum_{i\in [m]}||x_iQ - y_i||_2^2.
$$
Crucially, the orthogonality constraint on $Q$~\eqref{eq:orthoreg} ensures that the solution remains feasible, i.e. that $XQ \in \text{St}(n,k)$. Furthermore, we show in our experiments that this feasibility significantly improves generalization at very low label rates in comparison to standard Laplacian Eigenmaps SSL.
These interpretations serve to motivate our initialization and subsequent refinement. In particular, \citet{zhou2011error} consider the limiting behavior of Laplacian Eigenmaps SSL and show that it is non-degenerate in the limit of infinite unlabeled data.

\section{Iterative Algorithms for Graph-Based Semi-Supervised Learning on Stiefel Manifolds}

In this section, we introduce two iterative algorithms for optimization on Stiefel Manifolds. The first is a standard gradient projection method, whose convergence analysis under the Armijo rule is given in the appendix and referenced in the proof of convergence of the Sequential Quadratic Programming method (SQP). Next, we introduce Newton's method (or SQP). When the iterates are near a critical point, Newton's method is known to rapidly converge. Emulating Hager's algorithm, we will show that the Newton direction plays a roll in our SSM algorithm.


Recall that we consider quadratic minimization over the Stiefel manifold, problems of the following form:
\begin{equation}\label{eq:rescaled_f_apdx}
\begin{aligned}
\min_{X :X\in \textrm{St}(n,k)} \langle X, L X C \rangle - \langle X, B C^{1/2} \rangle
\end{aligned}
\end{equation}
Such problems are a generalization of well-known instances of nonconvex quadratic over the unit ball or sphere. These problems often arise in trust region methods \citep{sorensen82newtontrust, conn00trust}. Notably, there could exist many local solutions for \eqref{eq:rescaled_f_apdx}. We will demonstrate convergence to critical points of two iterative methods: a gradient projection method (Sec.~\ref{sec:pgd}) and the Sequential Subspace Method (Sec.~\ref{sec:ssm}) proposed in this work. 




The previous work on this topic has primarily focused on a specific case of \eqref{eq:rescaled_f_apdx}\textemdash i.e. \eqref{eq:rescaled_f_apdx} is one natural generalization of the constrained problem studied in \citep{Hager2001MinimizingAQ,Hager2005GlobalSSM}
\begin{equation}\label{eq:hager}
\min_x \{x^\top A x - 2\langle x, b\rangle: ||x|| = 1, x\in \mathbb{R}^n\}
\end{equation}
This problem is related to the trust region subproblem:
\begin{equation}\label{eq:trust}
\min_x \{x^\top A x - 2\langle x, b\rangle: ||x|| \leq 1, x\in \mathbb{R}^n\}
\end{equation}
The following two propositions describe the global solution of \eqref{eq:hager} (See \citep{sorensen82newtontrust}). The condition \eqref{eq:trust_opt_condition} states that the global solution $x$ is a critical point associated with $\lambda$ that is bounded above by the smallest eigenvalue $d_1$ of $A$. As we will discuss in Sec.~\ref{sec:globalconv}, these conditions may also be extended to optimization over the Stiefel manifold and also serve to motivate our active learning scheme. 
\begin{proposition}[\citet{Hager2005GlobalSSM}]
A vector $x \in \mathbb{R}^n$ is a global solution of \ref{eq:hager}, if and only if $||x|| = 1$, and
\begin{equation}\label{eq:trust_opt_condition}
A - \lambda I \succcurlyeq 0, \:\: (A - \lambda I)x = b
\end{equation}
holds for some $\lambda \in \mathbb{R}$.
\label{prop:hager1}
\end{proposition}
\begin{proposition}[\citet{Hager2001MinimizingAQ}]
Consider the eigenvector decomposition 
$$
A = [v_1, v_2, \ldots, v_n]\text{diag}(d_1, \ldots, d_n)[v_1, v_2, \ldots, v_n].T.
$$
Let $V_1$ be the matrix whose columns are eigenvectors of $\mathcal{L}$ with eigenvalue $d_1$. Then, $x = Vc$ is a solution for a vector $c$ chosen in the following way:
\begin{itemize}
\item Degenerate case: suppose $V_1^\top b = 0$ and $c_{\perp}:= ||(A - d_1I)^{\dagger}b||\leq 1$
Then, $\lambda = d_1$ and $x = (1 - c_\perp^2)^{1/2}v_1 + (A - d_1 I)^{\dagger} b$
\item Nondegenerate case: $\lambda < d_1$ is chosen so that $x = (A - \lambda I)^{-1}b$ with $||x|| = 1$.
\end{itemize}
\label{prop:hager2}
\end{proposition}
\begin{remark}[\citet{Hager2005GlobalSSM}]
Note that $||(A - \lambda I)^{-1}b||$ decreases monotonically with respect to $\lambda\leq d_1$. A proper value of $\lambda$ meets the condition $||x|| = 1$. A tighter bound on $\lambda$ can be estimated from 
$$
(d_1 - \lambda) ||V_1b||^2 \leq 1 = ||(A - \lambda I)^{-1}b||^2 \leq (d_1 - \lambda)^{-2}||b||^2
$$
With $||V_1 b|| > 0$, $\lambda$ lies in the interval $[d_1 - ||b||, d_1 - ||V_1 b||]$
\label{prop:hager3}
\end{remark}
It is clear from Proposition \ref{prop:hager2} that while a simple closed-form  solution to global solutions to sphere-constrained optimization is relatively easy to express, it depends on the complete diagonalization of the system matrix $A$, which is computationally prohibitive. As a consequence, much prior work has gone into the development of iterative algorithms which yield sequences of iterates that are convergent in the limit to solutions which satisfy certain optimality conditions. Here we will describe two well-known iterative methods for iterative quadratic optimization on the Stiefel Manifold.

Although the solutions to our more general problem are more challenging to characterize, we will later discuss that quadratic optimization problems on the Stiefel Manifold admit similar optimality conditions compared to quadratic problems on the sphere, i.e. the eigenvalues of the multiplier matrix $\Lambda$ are bounded by certain eigenvalues of the system matrix. Notably, we will prove that the sequential subspace method converges to a solution that satisfies this property. 

\subsection{Gradient Projection Method (PGD) Algorithm}
\label{sec:pgd}
We first introduce a projected gradient-based method. With appropriate step size $\alpha > 0$, PGD produces iterates $X_t$, $t = 1, 2, \ldots$
$$
X_{t+1} = [X_t - \alpha g_t]_+,
$$
where $g_t$ is given by the gradient of the objective of \eqref{eq:rescaled_f_apdx}\textemdash i.e. $g_t = LX_tC - BC^{1/2}$. Let $X_t' = X_t - \alpha g_t$. $[X_t']_+$ is the projection onto the manifold
$$
\mathcal{M} := \{X : X \in St(n,k), X^\top BC^{1/2} \geq 0\}.
$$
We first describe the projection $X = [X_t']_+$ as a composition of two projections; i.e. $[X_t']_+ = [[X_t']_{St}]_\mathcal{B} \in \mathcal{M}$:
\begin{flalign}
& Z = [X_t']_{St} := \argmin_Z \{||X_t' - Z||_F : Z \in St(m,r)\} \label{eq:stproj} \\
& X = [Z]_\mathcal{B}:= ZQ,\:\:Q=\argmin_Q \{||Z - BQ^\top||_F : Q \in O_k\} \label{eq:Bproj}
\end{flalign}
In other words, $Z \in St(n,r)$ and $Q \in O_k$ are chosen to minimize the sum
$$
||X_t' - Z||_F^2 + ||Z - BC^{1/2}Q^\top||_F^2 = ||X_t'Q - ZQ||^2_F + ||ZQ - BC^{1/2}||_F^2
$$
Take the SVD of $X_t'$, i.e. $X_t' = U_1D_1V_1^\top$. Then, the solution to \eqref{eq:stproj} is given by $Z = U_1V_1^\top$. Likewise, $X=ZQ$ for some orthogonal matrix $Q$ chosen to maximize $\langle X, BC^{1/2} \rangle = X^\top B$. 

\begin{proposition}[Projection onto $\mathcal{M}$]\label{prop:projection}
Consider the solution to the following projection:
\begin{equation}\label{eq:proj_apdx}
[X_t']_+ = \argmin_{X\in St(n,k)}\{\min_Q ||X - X_t'Q||_F^2 : X^\top B \geq 0, Q \in O_k\}
\end{equation}
Suppose the singular values of $X_t'$ and $B$ are positive. Then, the minimizer $X$ is uniquely determined by 
$$
X = [X_t']_+ = U_1 U_2 V_2^\top
$$
where $U_1, V_1, V_2$ are determined from the two SVDs,
$$
X_t' = U_1 \Sigma_1 V_1^\top, \:\:  U_1^\top BC^{1/2} = U_2\Sigma_2V_2^\top
$$
\end{proposition}
\emph{Proof.} The minimizer $X$ in \eqref{eq:proj_apdx} is the maximizer of $\max_X \langle X, X_t' Q\rangle$. Note that
$$
\langle X, X_t' Q\rangle = \langle X, U_1 \Sigma_1 V_1^\top Q \rangle = \langle U_1^\top X Q^\top V_1, \Sigma_1 \rangle \leq tr(\Sigma_1)
$$
Note two observations: (1) that $U_1^\top X Q^\top V_1$ lies in $O_k$, and (2) that equality holds if and only if $U_1^\top X Q^\top V_1 = I_k$ for some $XQ^\top V_1 \in \text{St}(n, k)$, i.e. $XQ^\top V_1 = U_1$, and thus,
$$
X = U_1 V_1^\top Q.
$$
Furthermore, the condition $X^\top BC^{1/2}$ is symmetric and positive definite implies a choice of $Q$ that fulfills 
$$
X^\top B = Q^\top V_1 U_2 \Sigma_2 V_2^\top, \:\: i.e., V^\top Q = V_2 U_2^\top
$$
Finally, note that since the singular values $\Sigma_1, \Sigma_2$ are distinct and positive, $U_1$ and $V_1$ are uniquely determined up to column-sign $Q_1 = \text{diag} (\pm 1, \ldots \pm 1)$. Likewise, $U_2$ and $V_2$ are uniquely determined up to $Q_2 = \text{diag} (\pm 1, \ldots \pm 1)$. Hence,
$$
X = U_1 Q_1 Q_2^\top U_2 Q_2 Q_2^\top V_2^\top = U_1 U_2 V_2^\top
$$
is unique. \qedsymbol

The convergence of the gradient method with Armijo rule is provided in the appendix.

\subsection{Sequential Quadratic Programming (SQP)}
\label{sec:sqp}

It is well-known that taking the Newton direction as the descent direction can speed up the convergence to a stationary point, particularly when the initialization is carefully chosen.
Following the principle of Sequential Quadratic Programming (SQP), we introduce the SQP direction $Z$ according to the linearization of \eqref{eq:rescaled_f_foc}, the first-order conditions of \eqref{eq:rescaled_f}:
\begin{align*}
(LZC - Z\Lambda) - X \Delta = E &:= BC^{1/2} - (LXC - X\Lambda), \quad
X^\top Z = 0
\end{align*}
\begin{proposition}[SQP iterate of the Lagrangian of \eqref{eq:rescaled_f_lagrangian}]\label{prop:sqp_iterate}
Assume $\Lambda$ is symmetric. Let $P^\perp = I - X^\top X$ be the projection onto the orthogonal complement of the column space of $X$ and $\Lambda C^{-1} = U\text{diag}([\lambda_1,\ldots,\lambda_k]) U^{-1}$ be the eigenvector decomposition of $\Lambda C^{-1}$. The Newton direction $Z$ of $X$ via the linearization of the first-order conditions is 
\begin{equation}
  Z = OU^\top,
  \label{eq:sqp}
\end{equation}
where each column of $O$, $o_j = (P^\perp LP^\perp - \lambda_j P^\perp)^\dagger BC^{-1}u_j$.
\label{prop:newton}
\end{proposition}
\emph{Proof.} Recall the FOC and its associated linearization with respect to descent directions of $X, \Lambda$; $(Z, \Delta)$:
\begin{align*}
&(LZC - Z\Lambda) - X \Delta = E := BC^{1/2} - (LXC - X\Lambda) \\
&X^\top Z = 0
\end{align*}
Applying the projection $P^{\perp} = I - XX^\top$ eliminates the $X\Delta$ term:
$$
PLZ - Z\Lambda C^{-1} = PLPZ - ZU\text{diag}([\lambda_1, \ldots, \lambda_k])U^{-1} = PEC^{-1}
$$
Equivalently, 
$$
PLPZU - ZU\text{diag}([\lambda_1, \ldots, \lambda_k]) = PEC^{-1}U.
$$ 
Let $O = ZU = [o_1,\ldots, o_k]$ lie in the range of $P$. Then, 
$$
PLo_j - \lambda_j = PEC^{-1}u_j, \text{ so } o_j = (PLP - \lambda_j P)^\dagger EC^{-1}u_j.
$$ \qedsymbol
\begin{algorithm}
\caption{SQP Update}
\begin{flushleft}
\textbf{Input:} System matrix $L$, affine term $B$, intermediate feasible iterate $X_t$, scaling term $C$\\
\textbf{Output:} $j-th$ columns of Newton update\textemdash $Z_j$
\end{flushleft}
\begin{algorithmic}[1]
\Function{SQP}{$L, \Lambda_t, B, X_t$}
\State $\Lambda_t = X_t^\top (LX_t C - BC^{1/2})$
\State $U\text{diag}([\lambda_1, \ldots \lambda_k])U = \Lambda C^{-1} = C^{-1/2}\Lambda_t C^{-1/2}$ 
\State init $O$, $P^\perp = I - X^\top X$
\For{$j \in [k]$}
    \State $o_j = (P^\perp L P^{\perp})^{\dagger}BC^{-1}u_j$
\EndFor
\State \Return $OU^\top$
\EndFunction
\end{algorithmic}
\label{alg:sqp}
\end{algorithm}

Algorithm~\ref{alg:sqp} presents the detailed steps involved in the computation of the Newton directions (Proposition~\ref{prop:newton}). In Section~\ref{sec:complexity}, we discuss its computational cost and in the following proposition, we demonstrate asymptotic convergence of the SQP method.
\begin{remark}
    The update $\Lambda_t \to \Lambda_{t+1}$ can be derived directly from $X$ via the least-squares estimate:
    \begin{equation}
        \min_{\Lambda} ||LXC - BC^{1/2} - X\Lambda||^2_F
    \end{equation}
    That is,
    \begin{equation}
        \Lambda = X^\top X \Lambda = X^\top (LXC - BC^{1/2}).
    \end{equation}
\end{remark}

\begin{remark}
    Convergence of the SQP method can be derived in a manner similar to that of PGD. The only difference is the computation of \eqref{eq:gd_convergence}. Using notation from Prop.~\ref{prop:pgd_convergence}, for any limit point $X' \in \mathcal{M}$, let $d'$ be the Newton direction $OU^\top$ and $P^\perp = I - X' X'^\top$. From Prop. \ref{prop:sqp_iterate},
    \begin{align}
        \langle \mathcal{P}(X'), d' \rangle &= \langle P^\perp (LX' - B), OU^\top \rangle \\
        &= \langle P^\perp (LX' - B)U, O \rangle \\
        &= \sum_{j=1}^r \langle P^\perp (LX' - B)u_j, (P^\perp L P^{\perp})^{\dagger}BC^{-1}u_j \rangle \\
        &= - \sum_{j=1}^r \langle P^\perp (LX' - B)u_j, (P^\perp L P^{\perp})^{\dagger}P^{\perp}(LX' - B)u_j \rangle \leq 0,
    \end{align}
    where we have used the fact that 
    \begin{equation}
        P^\perp E = P^\perp (B - (LX' - X' \Lambda)) = P^\perp (B - LX')
    \end{equation}
    and $P^\perp L P^\perp - \lambda_j P^\perp \succcurlyeq 0$. Note that the equality in eq. (29) holds if and only if $P^\perp(LX' - B)U = 0$. This completes the proof that any limit point is a stationary point. At any stationary point, we have $E = 0$ and thus $W = 0$ from (24). Finally, $Z = 0$, i.e., SQP terminates. 
\end{remark}
In the next section, we introduce the Sequential Subspace Method (SSM) on the Stiefel Manifold. Critically, SSM exhibits desirable convergence characteristics in comparison to the methods we presented in this section. In particular, SSM is guaranteed to produce iterates that decrease the objective value at saddle points and local maximizers (in contrast to gradient projection and Newton's method). Additionally, if the solution to the small-dimensional subproblem satisfies a certain second-order condition, in the limit SSM converges to a solution satisfying a certain second order necessary condition for optimality.

\section{Sequential Subspace Method (SSM)}
\label{sec:ssm}

In this section, motivated by the similarity between \eqref{eq:rescaled_f} and standard trust-region subproblems, we develop the framework of the \emph{Sequential Subspace Method (SSM)} on the Stiefel Manifold. In the $k=1$ and $C=I$ case, SSM has been applied to Trust-Region sub-problems with remarkable empirical results~\citep{Hager2001MinimizingAQ} and robust global convergence guarantees~\citep{Hager2005GlobalSSM}, even for so-called degenerate problems. SSM-based algorithms generate a sequence of iterates $X_t$ by solving a series of rescaled quadratic programs (of the same form as \eqref{eq:rescaled_f}) in subspaces of dimension much smaller than that of the original problem (where $d = |V|$, the number of vertices in the graph). Although stationary points can be recovered via generic iterative project-descent procedures (e.g., via SQP or trust-region-type algorithms), SSM is a computationally efficient algorithm designed to address scalability with respect to large problems. 

To give additional motivation for our method, we provide visualizations of predictions made by our proposed models in conjunction with Laplace learning. Note that a significant number of predictions made by Laplace Learning are concentrated around the origin. 
\begin{figure}[!ht]
\centering
\begin{subfigure}[b]{0.26\linewidth}
\includegraphics[width=\linewidth]{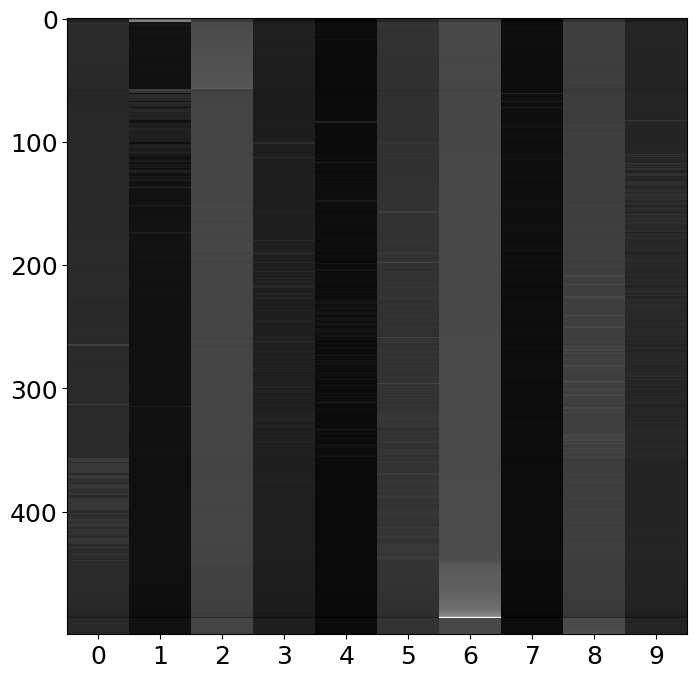}
\subcaption{}
\label{fig:a}
\end{subfigure}
\begin{subfigure}[b]{0.26\linewidth}
\includegraphics[width=\linewidth]{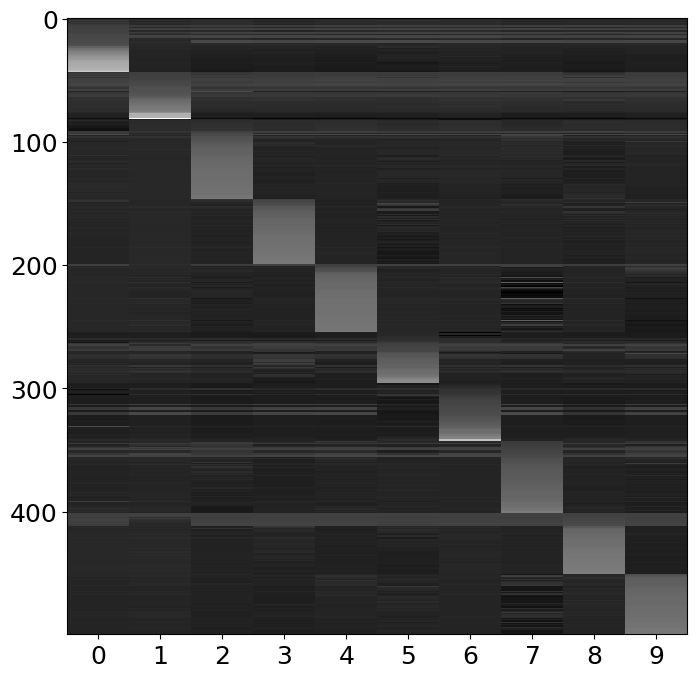}
\subcaption{}
\label{fig:c}
\end{subfigure}
\begin{subfigure}[b]{0.3\linewidth}
\includegraphics[width=\linewidth]{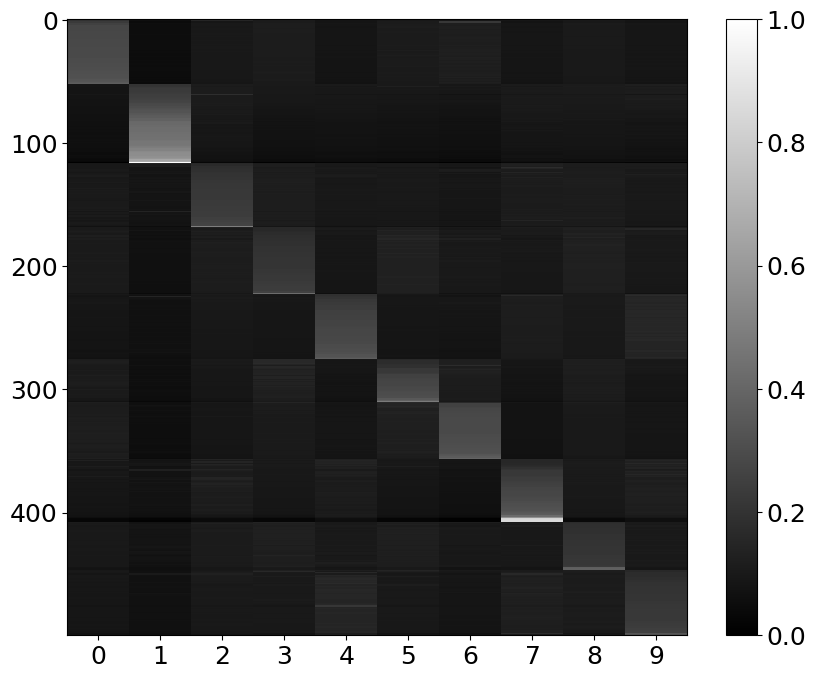}
\subcaption{}
\label{fig:e}
\end{subfigure}
\begin{subfigure}[b]{0.26\linewidth}
\includegraphics[width=\linewidth]{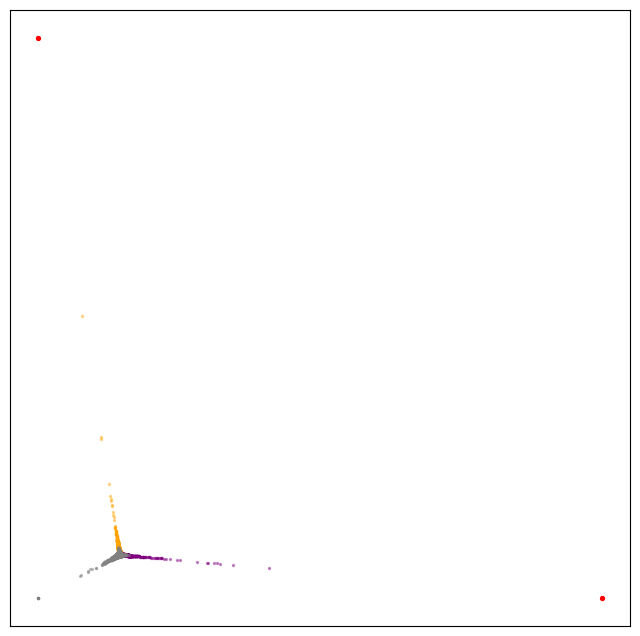}
\subcaption{}
\label{fig:b}
\end{subfigure}
%
%
\begin{subfigure}[b]{0.26\linewidth}
\includegraphics[width=\linewidth]{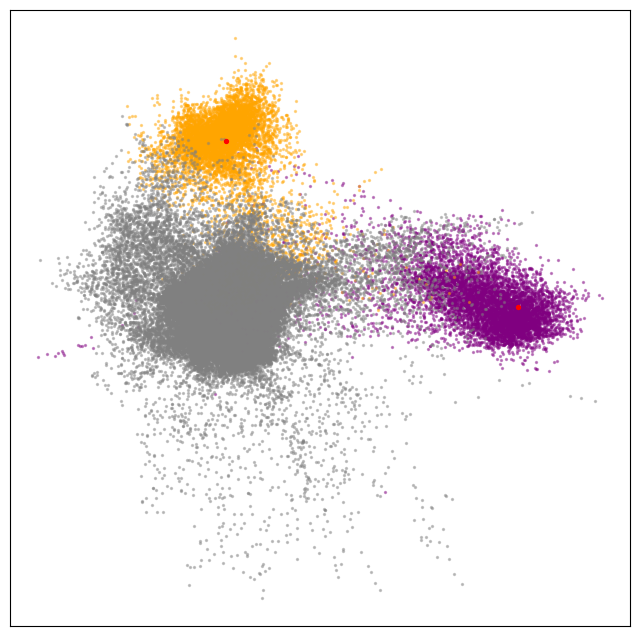}
\subcaption{}
\label{fig:d}
\end{subfigure}
%
%
\begin{subfigure}[b]{0.26\linewidth}
\includegraphics[width=\linewidth]{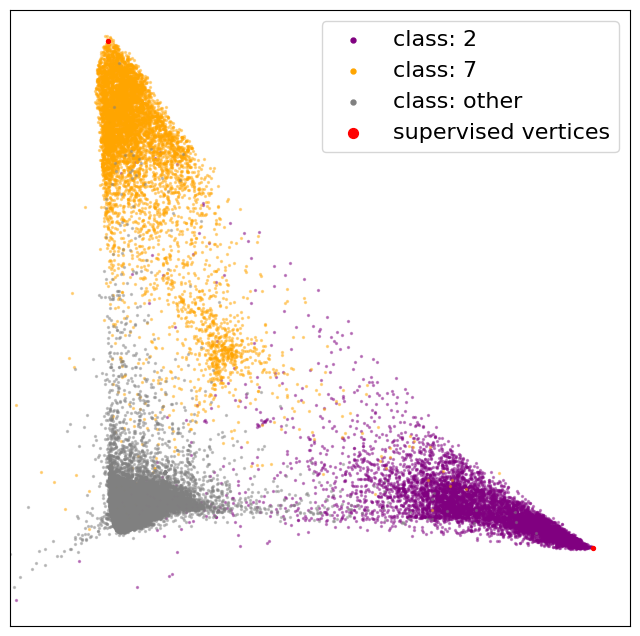}
\subcaption{}
\label{fig:f}
\end{subfigure}

%
\caption{Barcode plots of MNIST predictors (left) and embeddings of samples for digits `2' and `7' (right).
Learning is performed with 1 label per class. 
In the barcode plots, the rows are the samples, ordered by their class. Ordering of the columns was obtained by iteratively sorting the columns of the embedding matrices $X$. (\textbf{\subref{fig:a},\subref{fig:b}}) Laplace learning exhibits degeneracy in the limit of unlabeled data. (\textbf{\subref{fig:c},\subref{fig:d}}) Embeddings derived using Procrustes Analysis (Section~\ref{sec:approx}) exhibit no degeneracy but mixes samples from different classes together. (\textbf{\subref{fig:e},\subref{fig:f}}) SSM exhibits good classification performance (a block diagonally dominant barcode and well-separated embeddings) while respecting the geometry of unlabeled examples.}
\label{fig:mnist}
\end{figure}
In Figure \ref{fig:mnist}, we present 2-d visualizations of the embeddings of our SSM and Procrustes initialization method in conjunction with those produced by Laplace Learning. Each plot is constructed by taking the embedding (used to make predictions) implied by Laplace Learning, our approximate method, or SSM and the value associated with class ``2'' on one axis and ``7'' on the other axis. Ideally, there should be a clear and distinct cluster structure associated with classes 2 and 7 around the supervised points and the rest of the digits. Cluster structure should also be associated with the barcode plots via a block diagonally dominant barcode. The key message is that SSM exhibits a strong capability to discriminate classes (i.e. a block diagonally dominant barcode) while respecting the geometry of the unlabeled examples. In contrast, embeddings produced by Laplace Learning are not discriminatory (i.e. the barcode is uniform) and the embeddings are degenerate\textemdash concentrated at a single point.

To further adjust predictions, we introduce a multi-class Kernighan-Lin (KL) refinement algorithm to iteratively adjust the classification to improve a cut-based cost. Critically, this method is efficient (linear-time) and, in contrast to the gradient-based refinement method proposed in PoissonMBO~\citep{calder20poisson}, robust to the nonconvexity of the cut objective.


\subsection{SSM algorithm}
SSM works by solving a sequence of quadratic programs in subspaces of much smaller dimension relative to the size of the graph (the dimensions of $L$). SSM involves repeating the following pair of steps:
(1.) At step $t$ a tiny subspace, $S_t$, of dimension $4k \ll d$ is derived from the current iterate $X_t$, the gradient of the objective of \eqref{eq:rescaled_f} $g_t=LX_tC - BC^{1/2}$, an \emph{SQP} (i.e. Newton's method applied
to the first-order optimality system $X_t$) direction $Z_t$ derived in Prop.~\ref{prop:newton}, and the principal eigenvectors of $L$. The Sequential Quadratic Programming (SQP) framework~\citep{nocedal99sqp} is applied to compute $Z_t$. The computation of the SQP direction $Z_t = \text{SQP}(L,\Lambda, B, X_t,C)$ is given in Prop.~\ref{prop:sqp_iterate} and Alg.~\ref{alg:sqp}, line 5.  Let $V_t$ be the orthogonal matrix consisting of columns in $S_t$ (Alg.~\ref{alg:ssm}, lines 6 and 7), where
$$
S_t = \text{span}(X_t, Z_t, u, g_t).
$$
(2.) The next iterate $X_{t+1}$ is then generated by solving \eqref{eq:rescaled_f} in this small subspace. 
    $$
    [X_{t+1},\Lambda_{t+1}, u, \sigma] = \text{SSM}(L, B, S_t)
    $$
    consider the approximation $X = V_t \widetilde{X}$ for $\widetilde{X}$ given by
    \begin{equation}
    \min_{\widetilde{X} \in \text{St}(\widetilde{n},k)}F_{S} := \min_{\widetilde{X}}F(\widetilde{X}; V_t^\top L V_t, V_t^\top B).
    \label{eq:subspace_prob}
    \end{equation} Crucially, we highlight that when the eigenvectors of $L$ are included in the subspace, the sequence of iterates generated by SSM exhibits a global convergence property, which we discuss further in this section.

Note that \eqref{eq:subspace_prob} is solved using the Projected Gradient Method in practice.
Recall that $\Lambda_t$ according to the least-squares estimate derived from the first order condition in \eqref{eq:rescaled_f_foc} $\Lambda_t = X_t^\top (LX_tC - BC^{1/2})$.



%
\begin{algorithm}
\caption{Sequential Subspace Minimization on Stiefel Manifolds}
\begin{flushleft}
\textbf{Input:} System matrix $L$, eigenvectors $u$\\
\textbf{Output:} Embedding coordinates $X$
\end{flushleft}
\begin{algorithmic}[1]
\Function{SSM}{$L, u$}
\State Initialize $X$ according to Sec~\ref{sec:approx}.
\While{not converged} 
\State $Z \gets SQP(L, \Lambda, B, X, C)$ \Comment{Eq.~\ref{eq:sqp} \& Alg. 2}
\State $\mathcal{S} \gets \text{span}(X_t, Z_t, u, g_t)$
\State $V \gets QR(\text{col}(S))$
\State $L_t \gets V^\top L V$, $B_t \gets V^\top B$
\State $\widetilde{X} \gets \min_{X:X^\top X = I} F(X; L_t, B_t)$ \Comment{Solve Eq.~\ref{eq:rescaled_f} in $S$}

\State $X_t \gets V^\top \widetilde{X}$ \Comment{Lifted coordinates}
\State $t \gets t+1$
\EndWhile
\State \Return $X_t$ 
\EndFunction
\end{algorithmic}
\label{alg:ssm}
\end{algorithm}
\subsection{Analysis of SSM}

To show SSM converges, we first follow the proof of convergence of the Projected Gradient Method, Prop.~\ref{prop:pgd_convergence}\textemdash i.e. applying the Projected Gradient Method with step sizes chosen according to the Armijo rule ensures that any limit point $X_*$ is a stationary point, when $d_t = -(LX_tC - BC^{1/2}) \in S_t$. Let $V_t$ be an isometry, consisting of vectors in $S_t$ computed via a QR-factorization. Let $L_t := V^\top_t L V_t$ and $B_t = V_t^\top B$. Then, $F(\widetilde{X}; L_t, B_t)$ be the corresponding objective in $S_t$. SSM computes $X_{t+1} = V_t\widetilde{X}$, where
$$
\widetilde{X} := \argmin_{\widetilde{X}} F(\widetilde{X}; L_t, B_t)
$$
Note that the sequence $\{X_1,\ldots, X_t, \ldots\}$, with $X_{t+1}\in V_t$ monotonically reduces $F$ with respect to $t$:
\begin{flalign*}
F(X_{t+1}; L_t, B_t) &= \frac{1}{2}\langle X_{t+1}, LX_{t+1}C - 2B_tC^{1/2} \rangle \\
&\leq \min_{\widetilde{X}}\{\frac{1}{2}\langle V_t\widetilde{X}, LV_t\widetilde{X}C - 2BC^{1/2} \rangle = \frac{1}{2}\langle \widetilde{X}, L_t\widetilde{X}C - 2B_tC^{1/2} \rangle = F(\widetilde{X}; L_t, B_tC^{1/2})\} \\
&\leq \frac{1}{2}\langle X_t, LX_tC - 2BC^{1/2} \rangle = F(X_t; L, B)
\end{flalign*}
For each $t$, since the columns of $X_t$ and $LX_tC - BC^{1/2}$ lie in $S_t$, the iterations of the gradient projection method with Armijo rule lie in $S_t$, and the sequence with decreasing objective reaches a stationary point $\widetilde{X}$, it is ensured that the first order condition
$$
LX_*C - BC^{1/2} = X_*\Lambda_*
$$
holds for some matrix $\Lambda_* \in \mathbb{R}^{k\times k}$, given by
$$
\Lambda_* = X_*^\top (LX_*C - BC^{1/2}) = \lim_t X_t^\top(LX_kC - BC^{1/2})
$$
In the case $C = I$, the following states that the inclusion of $[u_1, \ldots, u_k]$ in $S_t$ improves the quality of the stationary point $X_*$, characterized by the eigenvalues of $\Lambda$.

\begin{proposition}[Eigenvalues of $\Lambda_*$]\label{prop:ssm_convergenceb}
Assume $C = I$. Let $X_* := [x_1,\ldots,x_k]$ be a stationary point generated from SSM. Then,
$$
LX_*C - X_*\Lambda_* = LX_* - X_*\Lambda_* = B.
$$
Let $\lambda_1,\ldots,\lambda_k$ be the eigenvalues of $\Lambda_*$ and let the eigenvalues of $L$ be $d_1 \leq d_2 \leq \ldots \leq d_n$. Then, $\max\{\lambda_1,\ldots,\lambda_k\} \leq d_k$. 
\end{proposition}

\noindent\emph{Proof. } Let $X_t = [x_{1,t}, \ldots, x_{k,t}]$ be a global minimizer in $V_{t-1}$. let $Y_t := [y_{1,t}, \ldots, y_{k,t}] = V_{t-1}^\top [x_{1, t}, \ldots, x_{k,t}]$. Then,
$$
L_{t-1}Y_t - B_{t-1} = Y_t \Lambda_t
$$
holds for some $\Lambda_k$ with eigenvalues $[\lambda_{1,t}, \ldots, \lambda_{k,t}]$. In addition, since $Y_tB_{t-1} = X_tB$, then $X_tB$ is positive semidefinite and symmetric. As $t \to \infty$, $X_*^\top B$ is also positive semidefinite and symmetric. Additionally, let
$$
P_j^{\perp, (t)} = I - \sum_{i \in \mathcal{R} - {j}} y_{i,t}y_{i,t}^\top.
$$
The second order condition implies
$$
P_j^{\perp, (t)}LP_j^{\perp, (t)} - \lambda_{j,t}P_j^{\perp, (t)} \succeq	0.
$$
Consider the optimality of $x_{1,t}$. Let $\phi_{1,t}$ by a unit vector orthogonal to $[x_{2,t},\ldots, x_{k,t}]$ in $span\{u_1,\ldots,u_k\}$. Then $P_j^{\perp, (t)}V_{t-1}^\top \phi_{1,t} = V_{t-1}^\top\phi_{1,t}$ holds and the second order condition yields
\begin{flalign*}
0&\leq \langle V_{t-1}^\top\phi_{1,t}, (P_j^{\perp, (t)}LP_{j,t}^{\perp} - \lambda_{j,t}P_{j,t}^{\perp})V_{t-1}^\top \phi_{1,t} \rangle \\
&= \langle \phi_{1,t}, (L - \lambda_{1,t}I)\phi_{1,t}\rangle\\
&\leq (\min_i d_i - \lambda_{1,t})||\phi_{1,t}||^2,
\end{flalign*}
Which implies $\lambda_{1,t} \leq \min_i d_i$. As $t \to \infty$, a subsequence of $\{x_{1,t}, \ldots, x_{k,t} : t \}$ converges to $[x_1,\ldots, x_k]\in \mathcal{M}$ and $\lambda_{1,t}$ converges to $\lambda_1$ Hence, $\lambda_1 \leq \min_i d_i$. Likewise, $\lambda_j \leq \min_i d_i$ by the optimality of $x_{j,t}$ for $j = 2,\ldots, k$. \qedsymbol

In the case $C \neq I$, the following proposition states that the inclusion of $[v_1, \ldots, v_k]$ in $S_t$ improves the quality of the stationary point $X_*$, characterized by the geometry of the optimization problem.

\begin{proposition}[Convergence of SSM]\label{prop:ssm_convergence}
Suppose $\bar{X}$ is a stationary point of \eqref{eq:cneqi}. For $S = \text{span}\{V, \bar{X}\}$, we have 
\begin{equation}
    \min_{X \in S \cup St(n,r)} F(X) \leq F(\bar{X})
\end{equation}
\end{proposition}

\noindent\emph{Proof. } For the stationary point $\bar{X}$, let $\bar{\Lambda}$ be the associated multiplier,
\begin{equation}
    F(X) = F(\bar{X}) + \frac{1}{2} \langle (X - \bar{X}), L (X - \bar{X}) C\rangle - \frac{1}{2} \langle (X - \bar{X}) \bar{\Lambda}, (X - \bar{X}) \rangle.
\end{equation}
Suppose $\bar{\Lambda}$ has at least one eigenvalue $\lambda_1$ larger than $d_k$. Then $\bar{X}$ is not a global minimizer. Let
\begin{equation}
    \bar{Y} = \bar{X}C^{1/2}U = [\bar{y}_1,\ldots, \bar{y}_k]
\end{equation}
Where $U$ are the orthogonormal eigenvectors of the matrix $C^{-1/2}\bar{\Lambda}C^{-1/2}$, i.e. let
\begin{equation}
U^\top C^{-1/2}\bar{\Lambda}C^{-1/2} U = \Gamma := \text{diag}(\gamma_1, \ldots, \gamma_k)
\end{equation}
and We can express $\bar{y}_1 = \sum_{j=1}^n \xi_jv_j$
for some scalars $\xi_j$.

Suppose $\xi_1 \neq 0$.  Take $Y = X^{1/2}U = [\bar{y}_1, \bar{y}_2,\ldots, \bar{y}_k]$ with
\begin{equation}
    y_1 = \bar{y}_1 - 2\xi_1 v_1 = - \xi_1 v_i + \sum_{j=2}^n \xi_j v_j.
\end{equation}
Note that $X\in St(n,r)$, since 
\begin{equation}
    X^\top X = (YU^\top C^{-1/2})^\top (YU^\top C^{1/2]}) = (\bar{Y}U^\top C^{-1/2})^\top (\bar{Y}U^\top C^{1/2]}) = \bar{X}^\top \bar{X} = I
\end{equation}
Then, we have that
\begin{equation}
    \begin{aligned}
        &F(X) - F(\bar{X}) = \frac{1}{2} \langle (X - \bar{X}), L (X - \bar{X}) C\rangle - \frac{1}{2} \langle (X - \bar{X}) \bar{\Lambda}, (X - \bar{X}) \rangle \\
        &= \frac{1}{2} \langle-2\xi_1v_1 e_1^\top, L(-2\xi_1v_1 e_1^\top) \rangle - \frac{1}{2} \langle (-2\xi_1v_1 e_1^\top)\Gamma, (-2\xi_1v_1 e_1^\top) \rangle \leq 2\xi_1^2 (d_k - \gamma_1) < 0.
    \end{aligned}
\end{equation}
\hfill \qedsymbol

The following is the result of Proposition 11.
\begin{theorem}[Global convergence of SSM]
\label{thm:ssm_convergence}
A limit $X_*$ of $\{X_1, X_2, \ldots, X_t,\ldots\}$ generated by SSM is a local minimizer of \eqref{eq:rescaled_f}. If $C = I$, 
SSM further satisfies the second-order condition $\max\{\lambda_1,\ldots,\lambda_k\} \leq d_k$ where $d_k$ is the $k$-th nonzero eigenvalue of $L$ and  $\lambda_1,\ldots,\lambda_k$ are the eigenvalues of $\Lambda_*$.
\end{theorem}

\subsection{On convergence to globally optimal solutions}
\label{sec:globalconv}

We briefly discuss the necessary conditions for \emph{global optimality} of \eqref{eq:rescaled_f}.

\begin{proposition}[Global solutions]
    Let $d_1$ be the smallest eigenvalue of $L$. Let $X'$ be a stationary point of 
    \begin{equation}
        \min_X F(X) \quad \text{s.t. }X^\top X = I
    \end{equation}
    and let $\Lambda'$ be the associated multipliers matrix. Suppose
    \begin{equation}
        d_1 C \succcurlyeq \Lambda'.
    \end{equation}
    Then $X'$ is a global minimizer. Suppose $d_1 C \succ \Lambda'$. Then, $X'$ is the unique global minimizer. 
\end{proposition}

\noindent\emph{Proof. } Let $\Lambda \in \mathbb{R}^{k\times k}$. The Lagrangian can be expressed:
\begin{equation} \label{eq:globalLang}
    \mathcal{G}(X) := \frac{1}{2} \langle X, LXC \rangle - \langle BC^{1/2}, X \rangle - \frac{1}{2} \langle \Lambda, X^\top X - I \rangle
\end{equation}
Consider any feasible $X \in St(n, k)$. Reformulate \eqref{eq:globalLang} in terms of a Taylor series of $X - X'$ around $X'$. Expanding yields
\begin{equation}\label{eq:goptcond}
    \begin{aligned}
        \mathcal{G}(X) &= \mathcal{G}(X') + \frac{1}{2} \{ \langle (X - X'), L(X - X')C \rangle - \rangle (X - X'), (X - X')\Lambda'\rangle\} \\
        &\geq \mathcal{G}(X') \frac{1}{2} \{\langle (X - X'), (X - X')(d_1 C - \Lambda')\rangle\} \geq F(X')
    \end{aligned}
\end{equation}
Since $\Lambda$ satisfies $d_1 C \geq \Lambda$, then \eqref{eq:goptcond} implies that
\begin{equation}
    F(X) = \mathcal{G}(X) \geq \mathcal{G}(X') = F(X')
\end{equation}
for each $X \in St(n,k)$, i.e., $X'$ is a global minimizer of $F$. On the other hand, suppose $F(X) = F(X')$ holds for some $X \in St(n,k)$. The condition $d_1 C \succ \Lambda$ implies $(X - X')^\top (X-X') = 0$, i.e., the uniqueness of $X'$. \hfill \qedsymbol

In general, this condition is restrictive. There is no guarantee that \emph{any} solution satisfies this condition. Alternatively, we may ensure recovery of a globally optimal solution if a non-degenerate condition on $BC^{1/2}$ is satisfied. Briefly, let $u_1, \ldots, u_k$ be the eigenvectors of $L$ corresponding to $k$ smallest nonzero eigenvalues $d_1 \leq \ldots \leq d_k$. At a high level, we need to ensure that the columns of $BC^{1/2}$ are not nearly orthogonal to $u_1, \ldots, u_k$. The following non-degeneracy condition on $BC^{1/2}$ ensures that any critical point $X$ satisfying a certain condition is a global minimizer, i.e. the projection of $BC^{1/2}$ on $U$ is sufficiently large, compared with the spectral gap $d_k - d_1$ for some $d_k$ such that $\Lambda \succcurlyeq d_k C$.

\begin{proposition}[Non-degeneracy condition]
Let $U = [u_1, u_2,\ldots, u_k]\in \mathbb{R}^{n \times r}$ be the eigenvectors of $L$ corresponding to the smallest $r$ nonzero eigenvalues $d_1 \leq d_2 \leq \ldots \leq d_k$. Let $X$ be a local solution satisfying the first order condition 
$$
LXC = X\Lambda + BC^{1/2}
$$
and second order condition $\lambda_1,\ldots, \lambda_k \leq d_k$. Let $s_1$ be the smallest singular value of $V^\top BC^{1/2} C^{-1} = V^\top BC^{-1/2}$. Suppose
\begin{equation}
   d_k - \gamma_j  \geq \sigma \text{ for all } j = 1, \ldots, r  
\end{equation}
and 
\begin{equation}
  \sigma > d_k - d_1
\end{equation}
Then, all eigenvalues $\gamma_1, \ldots, \gamma_k$ of the multiplier matrix $\Lambda C^{-1}$ are less than $d_1$ and $X$ is a global minimizer.
\end{proposition}
\emph{Proof. } Start with the first-order condition $LX = X\Lambda C^{-1} + BC^{1/2}C^{-1} = X\Lambda C^{-1} + BC^{-1/2}$. Let $v_j$ be a unit eigenvector of $\Lambda C^{-1}$ corresponding to eigenvalue $\gamma_j$. Taking the product of the first order condition with $u_j$ and $v_i$ yields
\begin{equation}
    d_i v_i^\top X v_j = u_i^\top LX v_j = (u_i^\top X) \Lambda C^{-1}v_j + u_i^\top BC^{-1/2}v_j,
\end{equation}
which implies 
\begin{equation}
    (d_i - \gamma_i)v_i^\top X u_j = v_i^\top BC^{-1/2}u_j.
\end{equation}
The second order condition indicates that 
\begin{equation}
    \gamma_j \leq d_k \text{ for } j = 1,\ldots, r
\end{equation}
Since $||V|| = 1 = ||X||$, then $||V^\top X u_j|| \leq ||Xu_j||\leq 1$. Since $|v_k ^\top BC^{-1/2} u_j|$ is bounded below by the smallest singular value of $V^\top BC^{-1/2}$, then with $i = r$, we have 
\begin{equation}
    |v_k^\top BC^{-1/2} u_j| \geq s_1
\end{equation}
and 
\begin{equation}
    d_k - \gamma_j \leq (d_k - \gamma_j)||v_k^\top X u_j|
\end{equation}
and since $(d_k - \gamma_j)||v_k^\top X u_j| = |v_k^\top BC^{-1/2} u_j|$, we have $d_k \geq \gamma_j$ for $j = 1,\ldots, r$. \hfill \qedsymbol

\label{sec:complexity}
\subsection{Complexity of SSM}
In this section, we discuss the computational cost of our method, dominated by the SQP routine
to compute the SQP directions. We claim that the per-iteration complexity of our algorithm is $T_{\text{matrix}}$, where $T_{\text{matrix}}$ is the complexity of each call to a sparse matrix (i.e. Laplacian, more generally an \emph{M-matrix}) solver. In particular, the QR-decomposition of $\text{col}(S)$ takes time linear in $n$. Likewise, fast, nearly linear-time solvers exist for solving Laplacian and Laplacian-like systems that are robust to ill-conditioning~\citep{spielmanteng}. We adopt Multigrid preconditioned conjugate gradient due to its empirical performance. We further note that the SSM procedure itself exhibits quadratic rates of convergence for nondegenerate problems and global convergence with \emph{at least} linear rates, even when the problem exhibits certain degenerate characteristics~\citep{Hager2005GlobalSSM}. 



\subsubsection{Computation of the descent direction $Z$}

In Sec.~\ref{sec:sqp}, we express the SQP direction $Z$ as the solution to the system characterized by the linearization of the first order optimality conditions. Namely, within each iteration of our procedure, we compute the Lagrangian multipliers as well as the SQP update for $X$ as defined in \eqref{eq:sqp}. As in Newton's method for unconstrained problems, SQP-based methods necessitate computation of inverse-vector products involving symmetric PSD matrices. 

We assume that by exploiting the sparsity of $L$, vector-vector and matrix-matrix multiplication can be done in linear time. In Alg.~\ref{alg:sqp}, we present the the SQP routine. The computation in line 3 involves an eigenvalue decomposition of a small $k\times k$ matrix. Thus, the primary overhead of our method lies in the computation of each column of $O$; $o_j, j = 1,\ldots, k$, which necessitates computation of $k$ Laplacian-like pseudoinverse-vector products.

\subsection{Cut-based refinement}
\label{sec:cutrefinement}
\begin{algorithm}
\caption{Kernighan-Lin refinement}
\begin{flushleft}
\textbf{Input:} KNN weights $W$\\
\textbf{Output:} Predictions $X$
\end{flushleft}
\begin{algorithmic}[1]
\State Compute $g(v)$ for all $v \in \mathcal{V}$ 
\While{not converged} 
\For{each pair of $n\choose 2$ partitions (classes) $\mathcal{V} = (\mathcal{V}_1, \mathcal{V}_2)$}
\State ordered list $l \gets \emptyset$ 
\State unmark all vertices $v\in V$ 
\For{$i=1$ to $n=\min(|\mathcal{V}_1|, |\mathcal{V}_2|)$}
\State $(v_1, v_2) \gets \argmax_{v_1, v_2} g(v_1, v_2)$
\State update $g$-values for all $v \in N(v_1) \cup N(v_2)$
\State add $(v_1, v_2)$ to $l$ and mark $v_1, v_2$
\EndFor
\State $k^* \gets \argmax_k \sum_{i=1}^kg(v_i, w_i)$ 
\State Update $(\mathcal{V}_1, \mathcal{V}_2)$: swap $(v_i, w_i) \in l$, $i = 1,\ldots,k^*$
\EndFor
\EndWhile

\end{algorithmic}
\label{alg:kl}
\end{algorithm}

Here we provide a detailed overview of the Kernighan-Lin (KL) algorithm and our multi-class extension.
The Kernighan-Lin algorithm \citep{klin1970} iteratively improves a given a disjoint bipartition of $\mathcal{V}$: $(\mathcal{V}_1, \mathcal{V}_2)$ such that $\mathcal{V}_1 \cup \mathcal{V}_2 = \mathcal{V}$, by finding subsets of each partition $A \subset \mathcal{V}_1$, $B \subset \mathcal{V}_2$ and then moving the nodes in $A$ and $B$ to the  opposite block. More concretely, the Kernighan-Lin algorithm repeatedly finds candidate sets $A$, $B$ to be exchanged until it reaches a local optimum with respect to the cut objective. Notably, the algorithm has the desirable tendency to escape poor local minima to a certain extent due to the way in which the sets $A$ and $B$ are created. This is one of the key features of the algorithm, and is a critical advantage over gradient-based methods for partitioning refinement, such as the MBO method presented in \citep{calder20poisson}.

The gain of a vertex $v$ is defined $g(v) = \sum_{j | \ell(v_i) = \ell(v_j)}W_{ij} - \sum_{j | \ell(v_i) \neq \ell(v_j)}W_{ij}$. , i.e. the reduction in the cut cost when the vertex $v$ is moved from partition $V_1$ to partition $V_2$. Thus, when $g(v) > 0$ we can decrease the cut by $g(v)$ by moving $v$ to the opposite block. Let $g(v,w)$ denote the gain of exchanging $v$ and $w$ between $V_1$ and $V_2$. Analogously, if $v$ and $w$ are not adjacent, then the gain is $g(v, w) = g(v) + g(w)$. If $v$ and $w$ are adjacent, then $g(v, w) = g(v) + g(w) - 2W_{vw}$.

The KL algorithm characterizes each vertex in $G$ as having one of two states: marked or unmarked. At each pass of the algorithm, each node is unmarked. A KL pass proceeds by iteratively finding an unmarked pair $v \in \mathcal{V}_1$ and $w \in \mathcal{V}_2$ for which $g(v, w)$ is maximum (note that $g(v, w)$ is not necessarily positive), marking $v$ and $w$, and updating the gain values of each of remaining unmarked nodes (i.e. the neighbors of $v$ and $w$) assuming an exchange between $v$ and $w$. This procedure repeats $p= \min(|\mathcal{V}_1|, |\mathcal{V}_2|)$ times.

After $p$ iterations, we have an ordered list $l$ of vertex pairs $(v_i , w_i)$, $i = 1,\ldots , p$. The swap-sets $A$ and $B$ are derived by finding the smallest index $k \in \{0,\ldots , p\}$ such that $P = \sum^k_{i=1} g(v_i, w_i)$ is maximum. Then, $A := \bigcup^k_{i=1}\{v_i\}$ and $B := \bigcup^k_{i=1}\{w_i\}$. A nonzero $k$ implies a reduction of the cut cost if $A$ and $B$ are exchanged. In this case, the exchange is performed and a new pass is instantiated. Otherwise, the KL iterations conclude.

Note that KL is typically performed over bi-partitions. We extend this framework to $k$-partitions in Algorithm~\ref{alg:kl} by considering a randomly ordered set of $k \choose 2$ pairs of classes (as defined by the predictions made on the vertex set) and performing KL on the subgraph restricted to these vertices. This procedure continues iteratively until all $k\choose 2$ pairs have been exhausted. Then, if convergence or a predetermined number of iterations has not been reached, a new random sequence is generated and the procedure continues.

\section{Graph-Based Active Learning}
\label{sec:graphal}
In this section, we introduce an active-learning scheme motivated by the criticality of label selection at low label rates and the benefits of diversity sampling. 
In the low label-rate regime, it is well-known that active learning strategies which emphasize \emph{exploration} of the sample-space, i.e. \emph{diversity} of the labeled samples, outperform those that rely on \emph{exploitation} of a classifier's decision boundary, e.g., notions of margin \citep{miller2022poisson}. Therefore, we propose a computationally efficient technique inspired by algebraic methods for selecting landmarks in graphs\textemdash i.e. a method that aims to select well-connected vertices diversely over the graph (vertices with large degree that are maximally separated) according to the spectral properties of the grounded Laplacian. This is further motivated by the discussion in Section \ref{sec:globalconv}, where we show that the spectral properties of the grounded Laplacian are intimately related to the convergence of SSM.

\subsection{Spectral score for diversity sampling on graphs}
\begin{figure}[ht!]
\centering
\begin{subfigure}[b]{0.24\linewidth}
\includegraphics[width=\linewidth]
{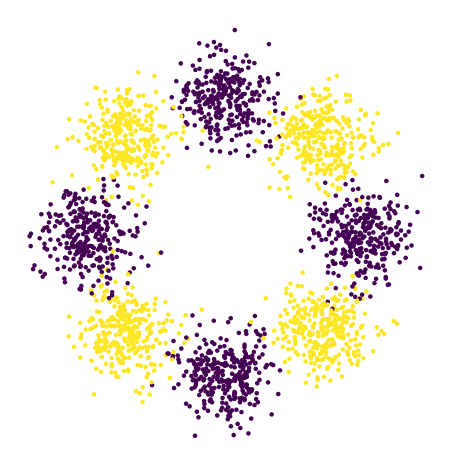}
\subcaption{class structure}
\end{subfigure}
\begin{subfigure}[b]{0.24\linewidth}
\includegraphics[width=\linewidth]
{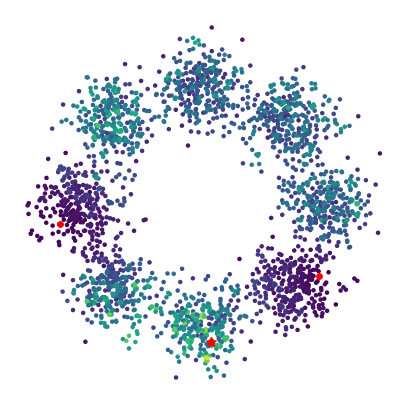}
\subcaption{iter 1}
\end{subfigure}
\begin{subfigure}[b]{0.24\linewidth}
\includegraphics[width=\linewidth]{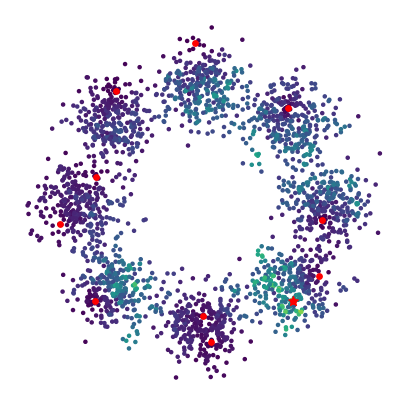}
\subcaption{iter 10}
\end{subfigure}
\begin{subfigure}[b]{0.24\linewidth}
\includegraphics[width=\linewidth]{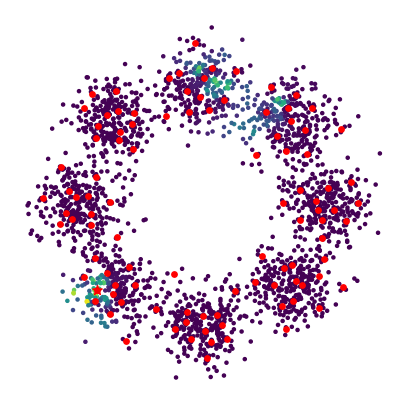}
\subcaption{iter 100}
\end{subfigure}
\caption{\textbf{Visualization of the lower-bound estimate on a ring of gaussians} Labeled points are annotated as red circles. Points to be labeled are marked as red stars. Brighter regions of the heatmap indicate vertices with higher score.}
\label{fig:visualization}
\end{figure}
We propose to select vertices from the set of unlabeled vertices according to the following measure:
\begin{equation}
\argmax_{v_i} \{s(v_i) := \widetilde{d}_{i}u_i^2\},
\end{equation}
where $\widetilde{d}_{i}$ denotes the degree of vertex $i$ defined on the sub-graph associated with the set of unlabeled vertices, $\widetilde{d}_i = \sum_{j \in \mathcal{U}}w_{ij}$ and $u_i$ corresponds to the $i$-th entry of $u$, the solution to the boundary-constrained eigenvalue problem:
\begin{equation}\label{eq:bce}
\left.\begin{aligned}
\mathcal{L} u_i &= \lambda u_i,&&\text{if }m+1 \leq i \leq M \\
u_i &=0,&&\text{if }1 \leq i \leq m \\
\end{aligned}\right\}.
\end{equation}
%
Note that $\text{supp}(u)$ is nothing but the entries of the eigenvector corresponding to the smallest eigenvalue of $\mathcal{L}_\mathcal{U}$. $u_i$ encodes various notions of centrality. Notably, \citet{cheng2019diffusiongeom} demonstrate an intimate connection between the solution $u$ in \eqref{eq:bv} for a normalized random walk Laplacian and the absorption time of a random walk, i.e. diffusion distance of vertex $i$, with respect to the boundary vertices $l$. More concretely, they prove that for solutions to boundary-constrained eigenvalue problems defined for certain Laplacians (e.g. absorbing random walk Laplacians), the diffusion distance from vertex $i$ to the boundary, $d_{l}(i)$ satisfies the following inequality:
$$
d_{l}(i) \log\left( \frac{1}{ |1 - \lambda_1|} \right) \geq \log \left( \frac{2|u(i)|}{||u||_{L^\infty}} \right)
$$
In other words, $d_l$ is \emph{highly correlated} with $|u|$. While \citet{cheng2019diffusiongeom} derive this relationship explicitly for $|u_i|$, we empirically show that selecting vertices for active learning in this way performs poorly relative to state of the art methods. Inspired by recent sampling strategies for graph signal reconstruction~\citep{ajinkya2018distancesampling} we expand on the analysis of \citet{cheng2019diffusiongeom} and show that in the presence of noise, \emph{reweighting} $u_i^2$ by $\widetilde{d}_i$ is an effective and principled heuristic. Additional details are provided in the supplemental material. Notably, we highlight that this score \emph{comes at no extra computational cost} due to certain features of SSM\textemdash in particular SSM's capability of producing estimates of the eigenvectors of $L$ in addition to solutions of \eqref{eq:rescaled_f}.

Intuitively, the proposed score naturally encodes the eigenvector centrality and degree of a vertex as well as its geodesic distance to labeled vertices. In practice, we incorporate eigenvectors of higher order eigenvalues:
\begin{equation}\label{eq:spectralscore}
s(v_i) = ||\widetilde{d}_{i} \odot U_i^2||_2,
\end{equation}
where $U$ is now an $n\times \ell$ matrix with eigenvectors of the grounded Laplacian as columns and $U_i^2$ denotes the matrix consisting of the square of the elements of column $U_i$. The choice of $\ell$ is left as a hyperparameter. In our experiments, we use $\ell=3$.

In other words, our score selects vertices \emph{that are both distant from the set  $\ell$ of labeled vertices, \textbf{and} well-connected}.
%
We provide an intuitive visualization of this score in Figure~\ref{fig:visualization}. The dataset is comprised of eight Gaussian clusters, each of equivalent size ($300$ samples), whose centers (i.e., means) lie evenly spaced apart on the unit circle. 
Each cluster is created by randomly sampling $300$ points from a Gaussian with mean $(\cos(\pi i/4),\sin(\pi i/4))^\top\in \mathbb{R}^2$ and fixed standard deviation $0.17$.  Classes are then assigned in an alternating fashion. 
For this example, efficient exploration via active learning is critical, particularly at low label rates. As we show in Figure~\ref{fig:visualization} our score facilitates effective exploration of the geometric clustering structure\textemdash i.e. by sampling diversely from each cluster in the ring.

\begin{remark}
If the set of labeled vertices, $l$, corresponds to the empty set, it is apparent that the smallest eigenvalue of $\mathcal{L}_\mathcal{U} = \mathcal{L}$ is $0$, and the corresponding eigenvector is $u=1$. Hence, the acquisition score of each vertex is nothing but a constant times its degree.
\end{remark}

\subsection{Random walk perspective}

While the work of \citet{cheng2019diffusiongeom} provides concrete motivation for our method, we derive the following property that ensures samples are diverse, i.e. far from the labeled nodes.
The component of the score involving the eigenvector of $\mathcal{L}_\mathcal{U}$ associated with the smallest eigenvalue is motivated primarily by previous work that investigates features encoded by the eigenvectors of the Laplacian of a graph with grounded vertices~\citep{cheng2019diffusiongeom}. These features specifically facilitate efficient methods to diversely sample the graph. Additionally, our score has connections to the graph signal processing literature~\citep{ajinkya2018distancesampling,anis215graphsampling} which aims to robustly recover a graph signal by sampling a sparse set of vertices. It has been demonstrated that one ideal sampling strategy that is robust to noise aims to maximize the smallest eigenvalue of the principal submatrix of the Laplacian, analogous to $\mathcal{L}_\mathcal{U}$. Below, we demonstrate that our method, namely squaring the entries of $u$ and weighting by $\widetilde{d}_i$, is directly related to one lower bound of the smallest eigenvalue of $\mathcal{L}_\mathcal{U}$.

Given a graph $G = (V, E, W)$ recall the graph Laplacian is defined to be $\mathcal{L} = D - A$. Consider a random walk over $V$ where the transition probabilities between two vertices $v_i$ and $vj$ are given by the entries of the degree-normalized edgeweights: $D^{-1}W$.
\begin{equation}
    P(V_k = v_j | V_{k-1} = v_i) = d_i^{-1}w_{ij}
\end{equation}
A state $v_i$ of a Markov chain is called absorbing if it cannot be exited, i.e., $Pr(X_{t+1} = v_i | X_t = v_i) = 1$.  The transition matrix has the following ``canonical'' form:
\begin{equation}
P = 
\begin{bmatrix}
Q & R\\
0 & I
\end{bmatrix},\:\:
P^t = 
\begin{bmatrix}
Q^t & \bar{R}\\
0 & I
\end{bmatrix}
\end{equation}
where $Q = D^{-1}W \in \mathbb{R}^{n \times n}$, $R$, and $\bar{R}$ are in $\mathbb{R}^{n \times m}$ are some nonzero matrices, $0$ is the zero matrix, and $I$ is an identity matrix. Intuitively, the first $n$ states are transient and the last $m$ states are absorbing. The probability of going to state $v_j$ from $v_i$ is given by $p_{ij}$. Furthermore $p^t_{ij}$ is the probability of being in state $v_j$ after $t$ steps when the chain is started in state $v_i$.

Define the fundamental matrix for $P$
\begin{equation}
N = \sum_{j=0}^\infty Q^j = (I - Q)^{-1}.
\end{equation}
The entry $n_{ij}$ of $N$ gives the expected amount of time a walker spends in $v_j$ if it starts from $v_i$. Likewise, the $i$-th entry of $N1$ is the expected number of steps before the walker is absorbed given that it starts in $v_i$.

We recall the following properties of a fundamental matrix for $P$.
\begin{definition}[Fundamental matrix]
For an absorbing Markov chain we define the fundamental matrix to be $N = (I-Q)^{-1} = \sum_{j=0}^\infty Q^j$.
\end{definition}

\begin{definition}[Arrival indicator $u^k$]
We define $n_i$ to be the total number of times that a random walker is in transient state $v_i$. $u^t$ is defined to be the vector-valued indicator with entries $u_i^t = 1$ if the process is in state $v_i$ after $t$ steps starting from any state, and $0$ otherwise.
\end{definition}

\begin{proposition}[Property of the fundamental matrix]
    Let $l$ be the set of absorbing (``labeled'') states and $p_{ij}^t$ be the probability of a random walker being in state $j$ after $t$ steps, starting from state $i$ i.e., $(\mathbb{E}[n_j])_i = N_{ij}$, where $v_i, v_j \notin l$. Note that $n_j = \sum_{t=0}^\infty u_j^t$. Then,
    \begin{equation}
    \begin{aligned}
        (\mathbb{E}[n_j])_i 
        = \sum_{k=0}^\infty Q^k
        = N _{ij}.
    \end{aligned}
    \end{equation}
\end{proposition}

\subsection{Computing central vertices}

To relate the absorbing walk to the graph Laplacian, let $v_s \in l \subset V$ denote an ``absorbing vertex'' and $\mathcal{L}_s$ denote the principal submatrix of $\mathcal{L}$, with the row and column associated with $v_s$ removed (labeled).

Consider an absorbing random walk over $V$, with the $s$-th vertex labeled corresponding to an absorbing state. Let $N_s\in \mathbb{R}^{n-1 \times n-1}$ denote the associated fundamental matrix, i.e. $(N_s)_{jk}$ denotes the expected number of visits to vertex $j$ from vertex $k$ before being absorbed. Then $N_s1$ denotes the expected number of steps before being absorbed by vertex $s$. 

\begin{proposition}{Upper-bound on the maximum expected commute time}
   Let $d_{\text{max}}$ denote the maximum degree of $G$ and $\{(u, \lambda_\text{min}(\mathcal{L}_s))\}_{i=1}^n$ denote the eigenpair corresponding to the smallest eigenvalue of the Laplacian submatrix $\mathcal{L}_s$, the smallest eigenvalue of $\mathcal{L}_s$ with the $i$-th vertex labeled. Denote the corresponding subgraph $G_s$ Let the scalar $u'_1 = \min_j (u / \max_i u_i)_j$ be the minimum entry of $u$ normalized such that its maximum element is $1$. Then, $\lambda_{\text{min}}(\mathcal{L}_s)$, satisfies the inequality
    \begin{equation}
        0 \leq \frac{u'_1}{d_{\text{max}}} \max_i[N_s 1]_i  \leq \frac{1}{\lambda_{\text{min}}(\mathcal{L}_s)},
    \end{equation}
    where $N_s$ corresponds to the fundamental matrix of an absorbing random walk on $G$. 
\end{proposition}
\noindent \emph{Proof. } 
First, note that $\mathcal{L}_s^{-1} = (D_s - W_s)^{-1} = (I - D_s^{-1}W_s)^{-1}D_s^{-1}$ is a symmetric, nonnegative matrix. Additionally, from the definition of the fundamental matrix (4), we have that $N_s = \mathcal{L}_s^{-1}D_s$. From Perron-Frobenius, we have that $u \geq 0$ elementwise. Let $u_1 = \min_j u_j \in \mathbb{R}_+$ and $u_2 = \max_j u_j \in \mathbb{R}_+$. Then,
%
\begin{equation}
    \begin{aligned}
        & 0\leq ||N_s1||_\infty = ||\mathcal{L}_s^{-1}D_s1||_\infty \leq d_{\text{max}}u_1^{-1}||\mathcal{L}_s^{-1}u||_\infty = \frac{1}{\lambda_{\text{min}(\mathcal{L}_s)} }\left (d_\text{max}\frac{u_2}{u_1}\right ) \\
    \end{aligned}
\end{equation}\hfill \qedsymbol

Following this justification, we compute a lower-bound on these eigenvalues in terms of $s$, based on Weyl's inequality, which characterizes the eigenvalues of a matrix under some additive perturbation.
\begin{proposition}[Lower bound on the Eigenvalues of the Laplacian submatrix]
    Let $\{(u_i, \lambda_i)\}_{i=1}^n$ be the ordered eigenpairs of the Laplacian submatrix $\mathcal{L}_s$. The smallest eigenvalue of the Laplacian submatrix $\mathcal{L}_s$ with the $i$-th vertex labeled, $\lambda'$, satisfies the inequality
    \begin{equation}
        \lambda' \geq \min_{j=1,\ldots,n}\{\lambda_j - \langle u^{(j)}, E^i u^{(j)} \rangle\}
    \end{equation}
\end{proposition}
\noindent\emph{Proof. }Consider a Laplacian submatrix $\mathcal{L}_s$ and the associated perturbation implied by labeling a vertex $E^i = \sum_{j \in \mathcal{U}}w_{ij} (E_{ij} + E_{ji})$, where $E_{ij}$ is an $n\times n$ matrix with $(\mathcal{L}_s)_{ij}$ in the $ij$-th position (and, to be clear, $w_{ij}$ is a scalar). Let $u$ be one eigenvector of $\mathcal{L}_s$. If $G_s$ is connected, the eigenvectors of $\mathcal{L}_s$ form a basis for $\mathbb{R}^n$.
Let $v$ be a unit eigenvector of $\mathcal{L}_s - E^i$ with eigenvector decomposition (where $u^{(j)}$ is the eigenvector associated with the $j$-th eigenvalue of $\mathcal{L}_s$):
$$
v = \sum_{j=1}^n t_j u^{(j)}
$$
for some coefficients $t_j$ s.t. $\sum_j t_j^2 = 1$. Then the eigenvalue $\lambda '$ of $\mathcal{L}_s - E^i$ is
\begin{equation}
\lambda' = v^\top (\mathcal{L}_s - E^i)v = \sum_{j=1}^n t_j^2 (\lambda_j - \langle u^{(j)}, E^i u^{(j)} \rangle) \geq \min_{j=1,\ldots,n}\{\lambda_j - \langle u^{(j)}, E^i u^{(j)} \rangle\}
\end{equation}
Thus, the maximum perturbation of the smallest eigenvalue of $\mathcal{L}_s - E^i$ is bounded below by the largest eigenvalue of $E^i$ (recall that $E^i$ has nonzero entries associated with the $i$-th column and $i$-th row of $\mathcal{L}_s$). \hfill \qedsymbol
\begin{remark}[Greedy maximization algorithm]
    Hence, to increase the eigenvalues of $\mathcal{L}_s$, a greedy selection implies the choice of $s$ that maximizes $-\langle u^{(j)}, E^i u^{(j)} \rangle= 2u^{(j)}_i \sum_{k \in \mathcal{U}} w_{ij} u^{(j)}_k$.
\end{remark}
\begin{remark}[Spectral gap]
When the spectral gap of $\mathcal{L}_s$ is large, $j$ need not be taken over $[n]$. More concretely, suppose $\lambda_{k' + 1} - \lambda_1 > \epsilon$, where $\epsilon$ is the largest eigenvalue of $\{E^i\:\: : \:\: i=1,\ldots, n\}$. Note that $\lambda' \geq \lambda$. Then,
\begin{equation}
\lambda' \geq \min \{\min_{j=1,\ldots,k'} \{ \lambda_j -  \langle u^{(j)}, E^i u^{(j)} \rangle \}, \lambda_{k' + 1} - \epsilon\} \geq \min_{j=1,\ldots,k'} \{\lambda_j - \langle u^{(j)}, E^i u^{(j)} \rangle\}
\end{equation}
\end{remark}

Algorithmically, on clustered graphs, the low-frequency eigenvectors (eigenvectors corresponding to the smallest eigenvalues of the Laplacian) are ``smooth'' over the graph and the score $\widetilde{d}_iu_i^2$ is a good proxy for the above bound, i.e.
\begin{equation}\label{eq:eig}
    2u^{(j)}_i \sum_{k \in \mathcal{U}} w_{ij} u^{(j)}_k \approx 2\widetilde{d}_{i}(u_i)^2.
\end{equation}
As we show below, using the ranking implied by \emph{just} $u_i$ corresponds to a diversity selection strategy that iteratively selects vertices that are far from the set of labeled nodes. Intuitively, weighting this measure by $\widetilde{d}_i$ encourages selection of vertices among those that are \emph{well connected}. Experimentally, as shown in the main text, this also has the effect of improving results.

%

\subsection{Summary of Algorithm and complexity of active learning}
In summary, our active learning framework repeats the following three steps:
\begin{enumerate}
    \item Apply SSM to derive $X_t^*$, the minimizer of $F(X; L_t, B_t, C_t)$ in \eqref{eq:cneqi}.
    \item Compute an estimate of the $k$ eigenvectors of $L_t = P\mathcal{L}_sP$ via the estimate $V_t\tilde{u}$, where $\tilde{u}$ are the eigenvectors of the small-dimensional SSM subproblem and compute the spectral score \eqref{eq:spectralscore}.
    \item Select the vertex with the largest score and query its label. Update problem parameters $L_{t+1}$, $B_{t+1}$, $C_{t+1}$.
\end{enumerate}

We now comment on the time and space complexity of graph-based active learning. In general, one would assume that the most expensive step is computing the principal eigenpairs of $\mathcal{L}_\mathcal{U}$. However, one key advantage of SSM is that it may provide accurate estimates of the principal eigenvectors of $L$, coinciding with the iterates $X_t$. In particular, 
$u = V \widetilde{u}$ is an estimate for the eigenvectors of $L$, if $\widetilde{u}$ consists of the eigenvectors of $L_t$ corresponding to the smallest $k$ eigenvalues. Thus, when iteratively deriving vertices to label via active learning and subsequently solving the graph-based SSL classification problem, we may effectively re-use the previous iteration's estimate of $u$ to do active learning in linear time, comparable to simple, decision-boundary-based margin methods and far more efficient compared to uncertainty uncertainty-based techniques that necessitate full or partial eigenvector decompositions of dense covariance matrices.

%
%
\section{Experiments}
\label{sec:experiments}

In this section, we present a numerical study of our algorithm applied to image classification in three domains at low label rates. We additionally explore medium and large label rates in comparison to recent state-of-the-art methods.

\subsection{Experimental setup}

We evaluated our method on three datasets: MNIST \cite{lecun1998gradient}, Fashion-MNIST \cite{xiao2017fashion} and CIFAR-10 \cite{krizhevsky2009learning}. As in \citet{calder20poisson}, we used pretrained autoencoders as feature extractors. For MNIST and Fashion-MNIST, we used variational autoencoders with 3 fully connected layers of sizes (784,400,20) and (784,400,30), respectively, followed by a symmetrically defined decoder. The autoencoder was trained for 100 epochs on each dataset. The autoencoder architecture, loss, and training are similar to \citet{kingma2013bayes}. 

For each dataset, we constructed a graph over the latent feature space. We used all available data to construct the graph, giving $n=70,000$ nodes for MNIST and Fashion-MNIST, and $n=60,000$ nodes for CIFAR-10. The graph was constructed as a $K$-nearest neighbor graph with Gaussian weights given by
\[w_{ij} =\exp\left( -4||x_i-x_j||^2/d_K(x_i)^2 \right),\] where $x_i$ represents the latent variables for image $i$, and $d_K(x_i)$ is the distance in the latent space between $x_i$ and its $K^{\rm th}$ nearest neighbor. We used $K=10$ in all experiments. The weight matrix was then symmetrized by replacing $W$ with $\frac{1}{2}(W+W^\top)$.


\subsection{Numerical results}
\begin{figure}[ht!]
\centering
\begin{subfigure}[b]{0.49\linewidth}
\includegraphics[width=\linewidth]{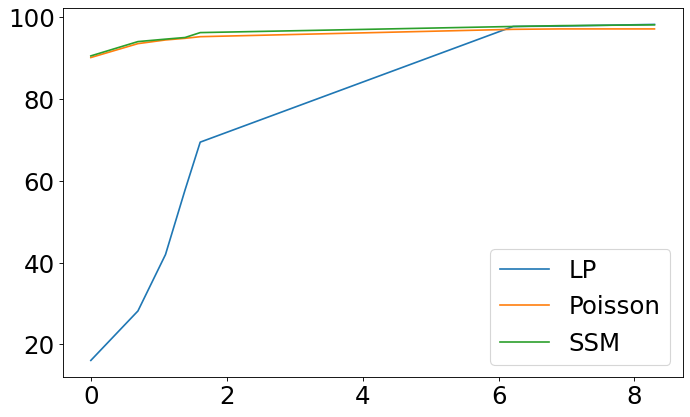}
\subcaption{}
\end{subfigure}
\begin{subfigure}[b]{0.48\linewidth}
\includegraphics[width=\linewidth]{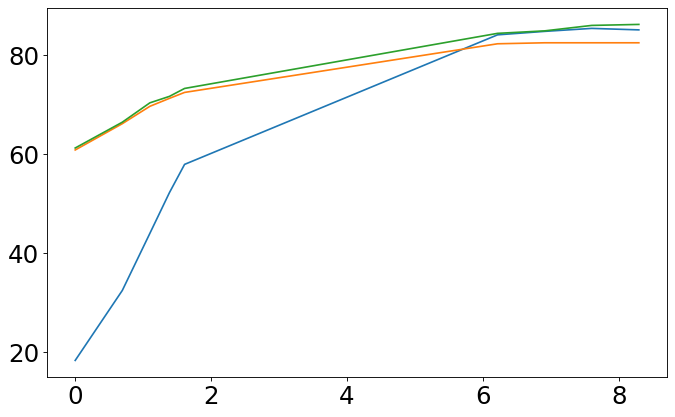}
\subcaption{}
\end{subfigure}
\caption{Scaling behavior as the number of labeled vertices increases beyond the low label rate regime the x-axis corresponds to the label rate ($\times 10^{3}$). the y-axis is accuracy. \textbf{(a)}: MNIST \textbf{(b)}: F-MNIST Average accuracy scores over 10 trials. We use the publicly available implementation of Poisson Learning~\cite{calder2017consistency}. }
\label{tab:scaling}
\end{figure}

\begin{table*}[!ht]
\caption{Average accuracy over 100 trials with standard deviation in brackets. Best is bolded.}
\label{tab:mainres}
\vskip 0.15in
\begin{small}
\begin{sc}
\begin{adjustbox}{width=\columnwidth,center}
\begin{tabular}{lllllll}
\toprule
\# FashionMNIST Labels per class&\textbf{1}&\textbf{2}&\textbf{3}&\textbf{4}&\textbf{5}&\textbf{4000}\\
\midrule
Laplace/LP \cite{zhu2003semi}&18.4 (7.3)      &32.5 (8.2)      &44.0 (8.6)      &52.2 (6.2)      &57.9 (6.7) &85.8  (0.0)      \\
Poisson \cite{calder20poisson}      &60.8 (4.6)      &66.1 (3.9)      &69.6 (2.6)      &71.2 (2.2)      &72.4 (2.3) & 81.1 (0.4)    \\
SSM       &  61.2 (5.3)   & 66.4 (4.1)   &  70.3 (2.3)   &  71.6 (2.0)    &  73.2 (2.1)  & 86.1 (0.1) \\
\cdashline{1-7}\noalign{\vskip 0.5ex}
Poisson-MBO \cite{calder20poisson}      &  62.0 (5.7)   &  67.2 (4.8)   &     70.4 (2.9) &   72.1 (2.5)  &  73.1 (2.7) &  86.8 (0.2)   \\
SSM-KL       &  \textbf{65.8 (1.1)}   &  \textbf{69.2 (1.2)}   &  \textbf{71.6 (1.2)}   & \textbf{73.0 (0.4)}     &  \textbf{73.4 (0.3)}  & \textbf{93.5 (0.1)}  \\
\midrule
\# CIFAR-10 & & & & & & \\
\midrule
Laplace/LP \cite{zhu2003semi}&10.4 (1.3)      &11.0 (2.1)      &11.6 (2.7)      &12.9 (3.9)      &14.1 (5.0)  &  80.9  (0.0)   \\
Poisson \cite{calder20poisson}      &40.7 (5.5)      & 46.5 (5.1)     & 49.9 (3.4)      & 52.3 (3.1)      & 53.8 (2.6) & 70.3 (0.9) \\
SSM       &  40.9 (6.1)   & 47.3 (5.9)   &  50.2 (4.3)    &  52.1 (4.3)    &  54.7 (3.4)  & 80.9 (0.1) \\[0.5ex]
\cdashline{1-7}\noalign{\vskip 0.5ex}
Poisson-MBO \cite{calder20poisson}      &41.8 (6.5)      & 50.2 (6.0)     &53.5 (4.4)      &56.5 (3.5)      & 57.9 (3.2) & 80.1 (0.3)  \\
SSM-KL       &  \textbf{43.7 (1.4)} &  \textbf{51.4 (1.3)}  &	\textbf{54.1 (2.1)} & \textbf{57.1 (1.3)} &	\textbf{58.8 (1.9)} &	\textbf{83.9 (0.0)} \\
\bottomrule
\end{tabular}
\end{adjustbox}
\end{sc}
\end{small}
\vskip -0.1in
\end{table*}
In table~\ref{tab:mainres}, we present our main results comparing our method to Laplace learning \citep{zhu2003semi} and Poisson learning \citep{calder20poisson} as well as our refinement based on KL-partitioning to the PoissonMBO refinement. 
Our SSM and SSM-KL methods consistently outperform state-of-the-art.
For a full evaluation, in Tables \ref{tab:MNIST}, \ref{tab:FashionMNIST}, \ref{tab:Cifar10} we compare our SSM approach and alignment-based approximation (Procrustes-SSL) against Laplace learning \citep{zhu2003semi}, Poisson learning \citep{calder20poisson}, lazy random walks \citep{zhou2004lazy,zhou2004learning},  weighted nonlocal Laplacian (WNLL) \citep{shi2017weighted}, $p$-Laplace  learning \citep{flores2019algorithms}, and Laplacian Eigenmaps SSL (LE-SSL)\citep{Belkin2002UsingMS}. 
Our SSM approach outperforms all methods in almost all cases.
Table \ref{tab:mainres} above and \ref{tab:MNIST}, \ref{tab:FashionMNIST}, \ref{tab:Cifar10}, and Figure \ref{tab:scaling}
show the average accuracy and standard deviation over all $100$ trials for various label rates. In particular, our method strictly improves over relevant methods on all datasets at a variety of label rates ranging from low (1 label) to high (4000). We further expand on this evaluation\textemdash showing that the trend persists with medium label rates (100-1000 labels).

On all datasets, the proposed method exceeds the performance of related methods, particularly as the difficulty of the classification problem increases (i.e. CIFAR-10).
In Tables~\ref{tab:MNIST}, \ref{tab:FashionMNIST}, and \ref{tab:Cifar10} we see that while Laplacian Eigenmaps SSL achieves better performance at higher label rates than Procrustes-SSL, Procrustes Analysis is significantly more accurate at lower label rates. 
We highlight the discrepancy between the approximate method (Procrustes-SSL) and our SSM-based refinement. This indicates the importance of SSM for recovering good critical points of \eqref{eq:rescaled_f}.

We compare our SSM approach and alignment-based approximation presented in section \ref{sec:approx} (Procrustes-SSL) against Laplace learning \cite{zhu2003semi}, Poisson learning \cite{calder20poisson}, lazy random walks \cite{zhou2004lazy,zhou2004learning}, weighted nonlocal Laplacian (WNLL) \cite{shi2017weighted}, $p$-Laplace  learning \cite{flores2019algorithms}, and Laplacian Eigenmaps SSL (LE-SSL)\cite{Belkin2002UsingMS}. 
In Tables \ref{tab:mainres}, \ref{tab:MNIST}, \ref{tab:FashionMNIST}, and \ref{tab:Cifar10} we restrict our comparison to methods \emph{without additional cut-based refinement} (e.g. PoissonMBO or KL), which we provide in Table~\ref{tab:mbo}. 

We conduct additional experiments to compare Procrustes-SSL + MBO, SSM + MBO, PoissonMBO, VolumeMBO at low ($1$, $3$, $5$) and high ($4000$) label-rates. Importantly, to conduct a fair comparison, we have augmented our proposed methods with the MBO-based refinement procedure proposed in Sec 2.4 of ~\citet{calder20poisson} with the same set of parameters. This amounts to replacing the PoissonLearning step (line 3, Algorithm 2 of \citet{calder20poisson}) with either of our proposed methods (Procrustes-SSL or SSM). We show that when our method is augmented with this additional refinement step, we gain significant improvements in solution quality as well as smaller standard deviations while outperforming all MBO-based approaches. This trend notably persists through the high-label-rate regime.

\begin{table*}[!ht]
\caption{MNIST: Average accuracy over 100 trials with standard deviation in brackets. Best is bolded.}
\label{tab:MNIST}
\vskip 0.15in
\begin{small}
\begin{sc}
\begin{adjustbox}{width=\columnwidth,center}
\begin{tabular}{llllll}
\toprule
\# Labels per class&\textbf{1}&\textbf{2}&\textbf{3}&\textbf{4}&\textbf{5}\\
\midrule
Laplace/LP \cite{zhu2003semi}&16.1 (6.2)      &28.2 (10.3)      &42.0 (12.4)      &57.8 (12.3)      &69.5 (12.2)      \\
Nearest Neighbor&55.8 (5.1)      &65.0 (3.2)      &68.9 (3.2)      &72.1 (2.8)      &74.1 (2.4)      \\
Random Walk \cite{zhou2004lazy}&66.4 (5.3)      &76.2 (3.3)      &80.0 (2.7)      &82.8 (2.3)      &84.5 (2.0)      \\
WNLL \cite{shi2017weighted}&55.8 (15.2)      &82.8 (7.6)      &90.5 (3.3)      &93.6 (1.5)      &94.6 (1.1)      \\
p-Laplace \cite{flores2019algorithms}&72.3 (9.1)      &86.5 (3.9)      &89.7 (1.6)      &90.3 (1.6)      &91.9 (1.0)      \\
Poisson \cite{calder20poisson}      &90.2 (4.0)      &93.6 (1.6)      &94.5 (1.1)      &94.9 (0.8)      &95.3 (0.7)      \\
LE-SSL \cite{Belkin2002UsingMS} &  43.1 (0.2)   &  87.4 (0.1)  &   88.2 (0.0)  &  90.5  (0.1)  & 93.7 (0.0)     \\
Procrustes-SSL &  87.0 (0.1)   &   89.1 (0.0)  &  89.1 (0.0)   &  89.6  (0.1)  &  91.4 (0.0)    \\
SSM       &   \textbf{90.6 (3.8)}   &    \textbf{94.1 (2.1)}  &   \textbf{94.7 (1.6)}  &    \textbf{95.1 (1.1)}  &  \textbf{96.3 (0.9)}   \\
\bottomrule
\end{tabular}
\end{adjustbox}
\end{sc}
\end{small}
\vskip -0.1in
\end{table*}
\begin{table*}[!ht]
\caption{FashionMNIST: Average accuracy scores over 100 trials with standard deviation in brackets. }
\label{tab:FashionMNIST}
\vskip 0.15in
\begin{center}
\begin{small}
\begin{sc}
\begin{adjustbox}{width=\columnwidth,center}
\begin{tabular}{llllll}
\toprule
\# Labels per class&\textbf{1}&\textbf{2}&\textbf{3}&\textbf{4}&\textbf{5}\\
\midrule
Laplace/LP \cite{zhu2003semi}&18.4 (7.3)      &32.5 (8.2)      &44.0 (8.6)      &52.2 (6.2)      &57.9 (6.7)      \\
Nearest Neighbor&44.5 (4.2)      &50.8 (3.5)      &54.6 (3.0)      &56.6 (2.5)      &58.3 (2.4)      \\
Random Walk \cite{zhou2004lazy}&49.0 (4.4)      &55.6 (3.8)      &59.4 (3.0)      &61.6 (2.5)      &63.4 (2.5)      \\
WNLL \cite{shi2017weighted}&44.6 (7.1)      &59.1 (4.7)      &64.7 (3.5)      &67.4 (3.3)      &70.0 (2.8)      \\
p-Laplace \cite{flores2019algorithms}&54.6 (4.0)      &57.4 (3.8)      &65.4 (2.8)      &68.0 (2.9)      &68.4 (0.5)      \\
Poisson \cite{calder20poisson}        &60.8 (4.6)      &66.1 (3.9)      &69.6 (2.6)      &71.2 (2.2)      &72.4 (2.3)      \\
LE-SSL \cite{Belkin2002UsingMS} &  22.0 (0.1)   &  51.3 (0.1)   &  62.0 (0.0)   &  65.4 (0.0)   &  63.2 (0.0)    \\
Procrustes-SSL &   50.1 (0.1)  &  55.6 (0.1)   &  62.0 (0.0)   &  63.4 (0.0)   &  61.3 (0.0)    \\
SSM       &  \textbf{61.2 (5.3)}    &   \textbf{66.4 (4.1)}  &  \textbf{70.3 (2.3)}   &    \textbf{71.6 (2.0)}  &  \textbf{73.2 (2.1)}    \\
\bottomrule
\toprule
\# Labels per class&\textbf{10}&\textbf{20}&\textbf{40}&\textbf{80}&\textbf{160}\\
\midrule
Laplace/LP \cite{zhu2003semi}&70.6 (3.1)      &76.5 (1.4)      &79.2 (0.7)      &80.9 (0.5)      & 82.3 (0.3)\\
Nearest Neighbor&62.9 (1.7)      &66.9 (1.1)      &70.0 (0.8)      &72.5 (0.6)      &74.7 (0.4)      \\
Random Walk \cite{zhou2004lazy}&68.2 (1.6)      &72.0 (1.0)      &75.0 (0.7)      &77.4 (0.5)      &79.5 (0.3)      \\
WNLL \cite{shi2017weighted}&74.4 (1.6)      &77.6 (1.1)      &79.4 (0.6)      &80.6 (0.4)      &81.5 (0.3)      \\
p-Laplace \cite{flores2019algorithms}&73.0 (0.9)      &76.2 (0.8)      &78.0 (0.3)      &79.7 (0.5)      &80.9 (0.3)      \\
Poisson \cite{calder20poisson}        &75.2 (1.5)      &77.3 (1.1)      &78.8 (0.7)      &79.9 (0.6)      &80.7 (0.5)      \\
LE-SSL \cite{Belkin2002UsingMS} &  67.1 (0.0)   &   68.8 (0.0)   &  70.5 (0.0)  &  70.9 (0.0)   & 66.6 (0.0)     \\
Procrustes-SSL &  65.3 (0.0)   &   66.2 (0.0)   &  68.3 (0.0)   &  69.6 (0.0)   &  64.5 (0.0)   \\
SSM       &  \textbf{76.4 (1.4)}    &   \textbf{78.1 (1.3)}   &  \textbf{79.4 (0.9)}    &   \textbf{80.3 (0.7)}   &    \textbf{82.6 (0.4)} \\
\bottomrule
\end{tabular}
\end{adjustbox}
\end{sc}
\end{small}
\end{center}
\vskip -0.1in
\end{table*}
\begin{table*}[!ht]
\caption{CIFAR-10: Average accuracy scores over 100 trials with standard deviation in brackets.}
\label{tab:Cifar10}
\vskip 0.15in
\begin{center}
\begin{small}
\begin{sc}
\begin{adjustbox}{width=\columnwidth,center}
\begin{tabular}{llllll}
\toprule
\# Labels per class&\textbf{1}&\textbf{2}&\textbf{3}&\textbf{4}&\textbf{5}\\
\midrule
Laplace/LP \cite{zhu2003semi}&10.4 (1.3)      &11.0 (2.1)      &11.6 (2.7)      &12.9 (3.9)      &14.1 (5.0)      \\
Nearest Neighbor&31.4 (4.2)      &35.3 (3.9)      &37.3 (2.8)      &39.0 (2.6)      &40.3 (2.3)      \\
Random Walk \cite{zhou2004lazy}&36.4 (4.9)      &42.0 (4.4)      &45.1 (3.3)      &47.5 (2.9)      &49.0 (2.6)      \\
WNLL \cite{shi2017weighted}&16.6 (5.2)      &26.2 (6.8)      &33.2 (7.0)      &39.0 (6.2)      &44.0 (5.5)      \\
p-Laplace \cite{flores2019algorithms}&26.0 (6.7)      &35.0 (5.4)      &42.1 (3.1)      &48.1 (2.6)      &49.7 (3.8)      \\
Poisson \cite{calder20poisson}        &40.7 (5.5)      &46.5 (5.1)      &49.9 (3.4)      &\textbf{52.3 (3.1)}      &53.8 (2.6)      \\ 
LE-SSL \cite{Belkin2002UsingMS} & 16.2 (0.1)   &  36.5 (0.1)   &   44.4 (0.1)  &  43.0 (0.0)   & 46.1 (0.0)     \\
Procrustes-SSL &  36.2 (0.1)   &  40.6 (0.1)   &  44.8 (0.1)   &  42.9 (0.0)   & 45.6 (0.0)     \\
SSM       &   \textbf{40.9 (6.1)}   &  \textbf{47.3 (5.9)}    & \textbf{50.2 (4.3)}     &  52.1 (4.3)    &  \textbf{54.7 (3.4)}    \\
\bottomrule
\toprule
\# Labels per class&\textbf{10}&\textbf{20}&\textbf{40}&\textbf{80}&\textbf{160}\\
\midrule
Laplace/LP \cite{zhu2003semi}&21.8 (7.4)      &38.6 (8.2)      &54.8 (4.4)      &62.7 (1.4)      &66.6 (0.7)      \\
Nearest Neighbor&43.3 (1.7)      &46.7 (1.2)      &49.9 (0.8)      &52.9 (0.6)      &55.5 (0.5)      \\
Random Walk \cite{zhou2004lazy}&53.9 (1.6)      &57.9 (1.1)      &61.7 (0.6)      &65.4 (0.5)      &68.0 (0.4)      \\
WNLL \cite{shi2017weighted}&54.0 (2.8)      &60.3 (1.6)      &64.2 (0.7)      &66.6 (0.6)      &68.2 (0.4)      \\
p-Laplace \cite{flores2019algorithms}&56.4 (1.8)      &60.4 (1.2)      &63.8 (0.6)      &66.3 (0.6)      & \textbf{68.7 (0.3)}      \\
Poisson \cite{calder20poisson}        &58.3 (1.7)      &61.5 (1.3)      &63.8 (0.8)      &65.6 (0.6)      &67.3 (0.4)      \\
LE-SSL \cite{Belkin2002UsingMS} &   47.9 (0.0) &  50.4 (0.0)  &  46.5 (0.0)  &   45.0 (0.0)  & 
46.7 (0.0)     \\
Procrustes-SSL &  46.1 (0.0)  &  50.0 (0.0)   &  46.9 (0.0)   &  45.5 (0.0)   &    46.9 (0.0)  \\
SSM       &   \textbf{59.4 (2.3)}   &  \textbf{62.4 (1.7)}    &   \textbf{64.9 (1.1)}   &    \textbf{66.6 (0.4)}  &  68.4 (0.4)   \\
\bottomrule
\end{tabular}
\end{adjustbox}
\end{sc}
\end{small}
\end{center}
\vskip -0.1in
\end{table*}

\begin{table*}[!ht]
\caption{Additional results comparing KL and MBO schemes at low and medium-high label rates.}
\label{tab:mbo}
\vskip 0.15in
\begin{center}
\begin{small}
\begin{sc}
\begin{adjustbox}{width=\columnwidth,center}
\begin{tabular}{lllll}
\toprule
MNIST \ \# Labels per class&\textbf{1}&\textbf{3}&\textbf{5}&\textbf{4000}\\
\midrule
PoissonMBO \cite{calder20poisson}&	96.5 (2.6)      &97.2 (0.1)      &97.2 (0.1)      &97.3 (0.0)        \\
VolumeMBO \cite{jacobs2018auction}&89.9 (7.3)      &96.2 (1.2)      &96.7 (0.6)      &96.9 (0.1)           \\
Procrustes-SSL + MBO &94.1 (0.1)      & 96.0 (0.0)      &97.1 (0.0)      &97.2 (0.0)      \\
SSM + MBO       &  \textbf{97.6 (0.1)}    &   \textbf{97.6 (0.1)}  &  \textbf{97.6 (0.1)}   &    \textbf{99.1 (0.0)}     \\
SSM-KL       &  \textbf{97.6 (0.1)}    &   \textbf{97.6 (0.1)}  &  \textbf{97.6 (0.1)}   &    \textbf{99.1 (0.0)}     \\
\bottomrule
\toprule
FashionMNIST \ \# Labels per class&\textbf{1}&\textbf{3}&\textbf{5}&\textbf{4000}\\
\midrule
PoissonMBO \cite{calder20poisson}&	62.0 (5.7) &	70.4 (2.9) &	73.1 (2.7) &	86.8 (0.2)        \\
VolumeMBO \cite{jacobs2018auction}&54.7 (5.2) &	66.1 (3.3) &	70.1 (7.1) &	85.5 (0.2)           \\
Procrustes-SSL + MBO &53.6  (2.8) &	60.3 (4.6) &	66.5 (3.2) &	70.1 (0.0)      \\
SSM + MBO       &  65.3 (1.9) &	71.4 (1.3) &	73.2 (0.6) &	93.4 (0.1)
     \\
     SSM + KL       &  \textbf{65.8 (1.1)}    &   \textbf{71.6 (1.2)}  &  \textbf{73.4 (0.3) }   &    \textbf{93.5 (0.1)}     \\
\bottomrule
\toprule
CIFAR-10 \ \# Labels per class&\textbf{1}&\textbf{3}&\textbf{5}&\textbf{4000}\\
\midrule
PoissonMBO \cite{calder20poisson}&	41.8 (6.5) &	53.5 (4.4) &	57.9 (3.2) &	80.1 (0.3)        \\
VolumeMBO \cite{jacobs2018auction}&38.0 (7.2) &	50.1 (5.7) &	55.3 (3.8) &	75.1 (0.2)           \\
Procrustes-SSL + MBO &38.1 (4.7) &	42.6 (3.3) &	46.4 (2.9) &	54.1 (0.1)      \\
SSM + MBO       &  42.3 (1.5) &	53.9 (2.5) &	57.9 (2.1) &	83.7 (0.0)
     \\
SSM-KL       &  \textbf{43.7 (1.4)}    &   \textbf{54.1 (2.1)}  &  \textbf{58.8 (1.9)} &	\textbf{83.9 (0.0)}     \\
\bottomrule
\end{tabular}
\end{adjustbox}
\end{sc}
\end{small}
\end{center}
\vskip -0.1in
\end{table*}

We additionally evaluate the scaling behavior of our method at intermediate and high label rates. In Figure~\ref{tab:scaling}, we compare our method to Laplace learning and Poisson learning on MNIST and Fashion-MNIST with 500, 1000, 2000, and 4000 labels per class. We see significant degradation in the performance of Poisson learning, however, our method maintains high-quality predictions in conjunction with Laplace learning. These results imply that while Laplace learning suffers degeneracy at low label rates and Poisson learning seemingly degrades at large label rates, our framework performs reliably in both regimes\textemdash covering the spectrum of low and high supervised sampling rates. 

\subsection{Comparison with an open-source tool for Riemannian optimization}
In this section, we highlight the practical efficacy of SSM by comparing to existing standard open-source implementations~\cite{pymanopt} of benchmark optimization algorithms~\cite{absiltrust}.
\begin{table}[!ht]
\caption{Tool comparison: wall time per-iteration, \# iterations to reach |grad| <= 10e-5 (-- denotes no convergence), and accuracy using MNIST digits restricted to 0-5 with 1 label / class.}
\label{tab:runtime}
\begin{center}
\begin{small}
\begin{sc}
\begin{tabular}{lllll}
\toprule
  & wall time / iter & \# iter to crit. point &  accuracy & \\
\midrule
SSM & 6.1 & 7 & 0.99 & \\
TR & 145.5 & 10 & 0.94 & \\
RG & 3.4  & -- & 0.75 & \\
\bottomrule
\end{tabular}
\end{sc}
\end{small}
\end{center}
\end{table}

\begin{figure}[ht!]
\centering
\begin{subfigure}[b]{0.49\linewidth}
\includegraphics[width=\linewidth]{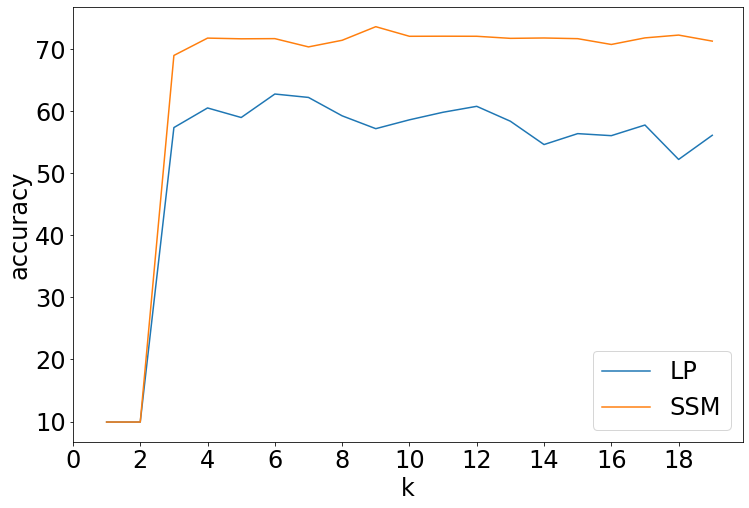}
\subcaption{}
\label{fig:knn}
\end{subfigure}
\begin{subfigure}[b]{0.49\linewidth}
\includegraphics[width=\linewidth]{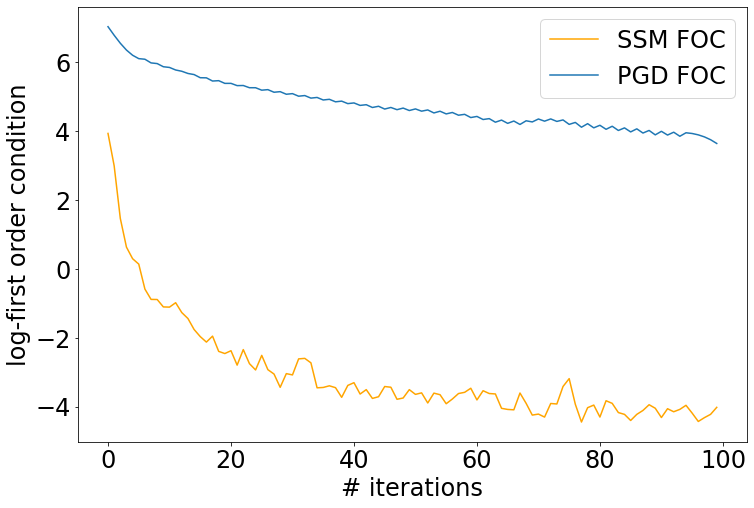}
\subcaption{}
\label{fig:foc}
\end{subfigure}
\caption{\textbf{Robust performance of SSM on F-MNIST}. (\textbf{\subref{fig:knn}}) robustness to different numbers of neighbors $k$ used to construct the graph, averaged over $10$ trials, 5 labels per-class. (\textbf{\subref{fig:foc}}) The log-first order condition, i.e. empirical rate of convergence of Projected gradient method and SSM on F-MNIST with 5 labels per-class. }
\label{fig:layout}
\end{figure}
We include a comparison between the SSM component of our framework and the general first (RG) and second-order (TR) Riemannian optimization algorithms implemented in the Pymanopt package~\cite{pymanopt}. For submanifolds of Euclidean spaces, first-order methods for constrained problems that consist of iteratively taking a tangent step in the Euclidean space followed by a projection (i.e. our projected gradient method) are functionally equivalent to Riemannian Gradient methods~\cite{absil07matrixmanopt, absil12projectionretraction}. In our work, we consider the Euclidean Projection onto the Stiefel manifold given by the SVD. However, Pymanopt, by default, defines the retraction via the QR decomposition (although SVD-based retractions are also supported). In theory, this should yield similar convergence results in theory and practice.

We also consider the second-order trust-region method supported by Pymanopt~\cite{absiltrust}. We note that one contribution of our work is the extension of SSM, a state-of-the-art method for large-scale trust-region subproblems to more general QCQPs. More generally, our generalization and application of SSM is motivated by its success on large-scale trust-region subproblems (which our problem shares many characteristics with) - with remarkable empirical results and robust convergence guarantees, even for so-called ``degenerate problems''. SSM has many advantages over typical methods for Riemannian optimization (including first and second-order trust-region methods). To summarize, we claim that our proposed SSM method + Procrustes initialization, designed specifically for the problem we propose should outperform the more general techniques implemented in Pymanopt.

We show in Fig.~\ref{fig:foc} that the choice of subspaces plays a critical role in the rate and quality of convergence of our method (compared to first-order methods). One may also ask how our method compares to traditional second-order methods (e.g. Riemannian Trust-region). In theory, SSM employs a special set of vectors to estimate the Hessian information to update the search direction via subspace minimization. As a result, the Hessian information estimated from SSM is usually better than the Hessian estimated from CG or BFGS methods typically used for trust-region type approaches.

Figure \ref{fig:knn} shows the accuracy of SSM at 5 labels per class as a function of the number of neighbors $K$ used in constructing the graph, showing that the algorithm is not significantly sensitive to this choice. 

%
%
In Figure \ref{fig:foc}, we demonstrate the convergence behavior of SSM and the projected gradient method discussed previously by plotting the norm of the first order condition (FOC): $||LX_tC - BC^{1/2}-X_t\Lambda_t||$. Note that while both methods are guaranteed to monotonically reduce the objective of \eqref{eq:rescaled_f} via line search, SSM rapidly converges to a critical point, while the projected gradient method fails to converge, even after 100 iterations.

\subsection{Spectral algorithm for active learning}
\begin{figure}[ht!]
\centering
\includegraphics[width=\linewidth]{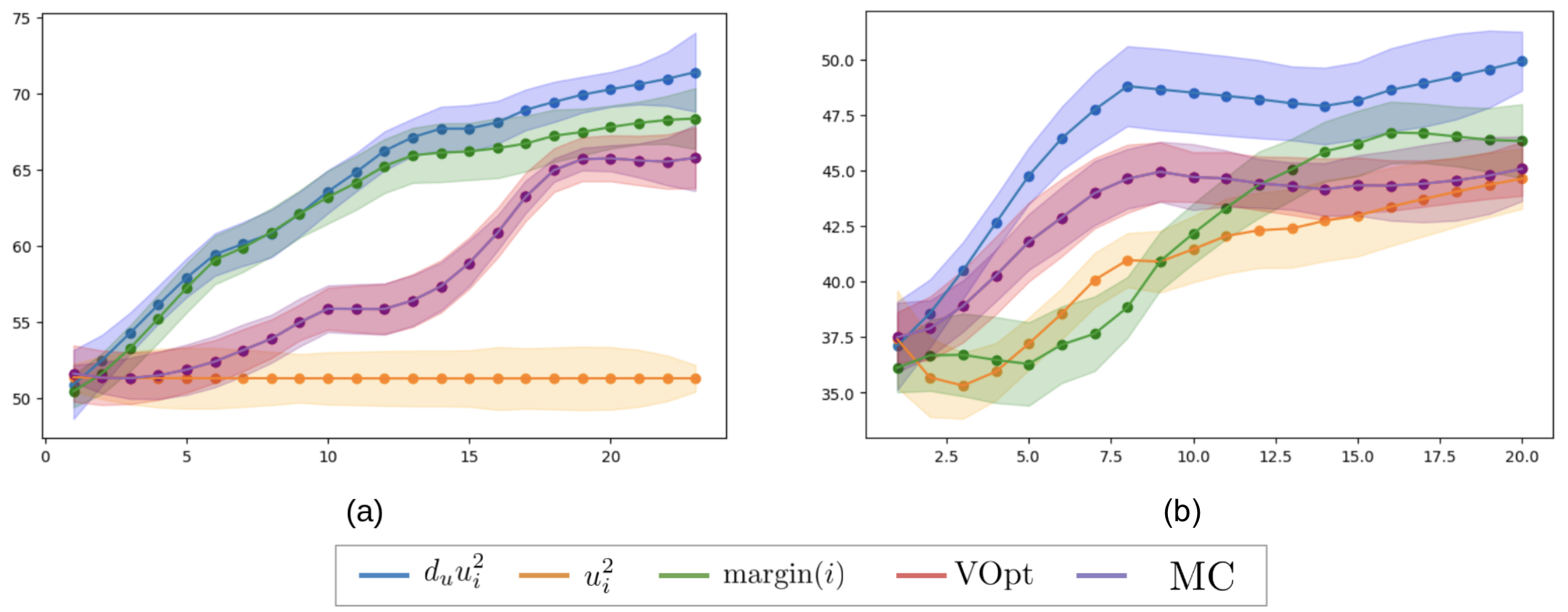}
\caption{\textbf{Performance of SSM with active learning on F-MNIST (a)  and CIFAR-10 (b)} Comparison between active learning methods using SSM-KL. x-axis denotes the number of vertices of the graph queries. y-axis denotes the accuracy over 10-trials (initial labeled set). The shaded region denotes $0.5\sigma$.
}
\label{fig:active_learning}
\end{figure}

We numerically evaluate our selection scheme for active learning on FashionMNIST and CIFAR-10 in Figure \ref{fig:active_learning}. We compare to minimum margin-based uncertainty sampling~\cite{burractivelearning2012}, VOpt~\cite{jivopt2012}, and Model Change (MC)~\cite{miller2021modelchange}. Note that uncertainty sampling selects query points according to the following notion of margin:
$
\text{margin}(i) = \argmax_j (X_i)_j - \argmax_{k \neq j} (X_i)_k.
$
One can interpret a smaller margin at a node as more uncertainty in the classification. We additionally note that MC and VOpt necessitate eigendecompositions of certain covariance matrices. Our score is implemented as
$$
s'(v_i) = s(v_i) - \lambda_t\cdot \text{margin}(X),
$$
where $\lambda$ increases with $t$ via $\lambda_{t+1} = \left(1 + \epsilon^{1/2k}\right)\lambda_{t}$ for some small value of $\epsilon=10^{-4}$.
%
%
%
We show that when coupled with the proposed SSM algorithm in an iterative fashion our active learning scheme outperforms related methods at low-label rates across all benchmarks. We also emphasize that due to certain features of SSM, the computation of $u_i$ is obtained for free after the first iteration. 
\section{Conclusion}
\label{sec:conclusion}
We have proposed a novel formulation of semi supervised and active graph-based learning. Motivated by the robustness of semi-supervised Laplacian eigenmaps and spectral cuts in low label rate regimes, we introduced a formulation of Laplacian Eigenmaps with label constraints as a nonconvex Quadratically Constrained Quadratic Program. We have presented an approximate method as well as a generalization of a Sequential Subspace Method on the Stiefel Manifold. In a comprehensive numerical study on three image datasets, we have demonstrated that our approach consistently outperforms relevant methods with respect to semi-supervised accuracy in low, medium, and high label rate settings. We additionally demonstrate that selection of labeled vertices at low-label rates is critical. 
An active learning scheme is naturally derived from our formulation and we demonstrate it significantly improves performance, compared to competing methods. Future work includes a more rigorous analysis of the active learning score and of the problem in \eqref{eq:rescaled_f} and our algorithmic generalization of SSM\textemdash for example, conditions on $L$ and $\mathcal{U}$ that guarantee convergence to globally optimal solutions with convergence rates derived in \citet{Hager2001MinimizingAQ,Hager2005GlobalSSM, absiltrust}. 


\acks{This work is partially supported by NSF-CCF-2217058.}


\newpage

\appendix


\section{Additional Proofs}

\subsection{Convergence of the Projected Gradient Method (PGD)}
The convergence of the gradient method with Armijo rule is provided in the following proposition. The step size $\alpha$ is selected to improve the objective function via Armijo's rule.

\begin{remark}\label{rem:differential}
Consider the function $h(X) = [X]_{\text{St}}$ defined on $\mathbb{R}^{n\times k}$ and $X\in St(n,k)$. The differential $\mathbb{D}h$ at $X$ is the linear map given by
$$
\mathbb{D}h(X)[T] - \lim_{\alpha \to 0}\alpha^{-1}(h(X + \alpha T)-h(X)) = (I - XX^\top)T + (-1/2)X(T^\top X - X^\top T)
$$
for each $T\in \mathbb{R}^{n\times k}$. When $X \in \mathcal{M}$ and $T = -(LXC - BC^{1/2})$, then
$$
\langle T, \mathbb{D}h(X)[T] \rangle = ||(I- XX^\top)T||_F^2.
$$
\end{remark}

\begin{proposition}\label{prop:pgd_convergence}
Let $d_t = -(LX_tC - BC^{1/2})$. Let $\{X_t\}$ be a sequence generated by the gradient projection method
$$
X_{t+1} = [x_t + \alpha_t d_t]_+
$$
where $\alpha_t$ is chosen according the Armijo rule. Then, every limit point of $\{X_t\}$ is a stationary point.
\end{proposition}
\emph{Proof.} The proof is motivated by the proof by contradiction of Prop. 1.2.1 in \citet{bertsekas99nonlinprog}.

Let $\mathcal{P}(X) = (I - X^\top X)(AXC - BC^{1/2})$ be the projected gradient of the objective of \ref{eq:rescaled_f_apdx}, $F$ at $X$. We define $\alpha$ given by the Armijo rule\textemdash i.e. let $s > 0$, $\sigma \in (0,1)$ and $\beta \in (0,1)$. $\alpha_t = \beta^{m_t}s$, where $m_t$ is the first nonnegative integer $m$ for which
$$
F(X_t) - F([X_t + \beta^msd_t]_+)\geq -\sigma \beta_t s\langle \mathcal{P}(X_t), d_t \rangle
$$
Suppose $\hat{X}\in \mathcal{M}$ is a limit point of $\{X_t\}$ with $||\mathcal{P}(\hat{X}|| > 0$. By definition, $\{F(X_t)\}$  is monotonically nonincreasing to $F(\hat{X})$., i.e. $F(X_t) - f(X_{t-1}) \to 0$. By definition, since the $\alpha_t$, the step sizes are generated via the Armijo rule, $a_t$ satisfies
\begin{equation}\label{eq:armijo}
\begin{aligned}
F(X_t) - F(x_{t+1}) &\geq F(X_t) - F([X_t + \alpha_t d_t]_+) \\
&\geq -\sigma \langle \mathcal{P}(X_t), d_t \rangle = \sigma \alpha_t ||\mathcal{P}(X_t)||_F^2
\end{aligned}
\end{equation}
Let $\{X_t\}_{\mathcal{T}}$ be a subsequence converging to $\hat{X}\in \mathcal{M}$ Since 
$$
\lim_{t \to \infty} \sup -\langle \mathcal{P}(X_t), d_t \rangle = ||\mathcal{P}(\hat{X})||^2 > 0,
$$
\eqref{eq:armijo} implies $\{\alpha_t\}_{\mathcal{T}}\to 0$. From Armijo's rule, for some $t'\geq 0$, the inequality
\begin{equation}\label{eq:armijo2}
F(X_t) - F([X_t + \alpha_t \beta^{-1}d_t]_+)< -\sigma \alpha_t \beta^{-1} \langle \mathcal{P}(X_t), d_t \rangle
\end{equation}
holds for all $t \geq t'$. By taking a subsequence $\{d_t\}_{\mathcal{T}'}$ of  $\{d_t\}_{\mathcal{T}}$ such that $\{d_t\}_{\mathcal{T}'} \to d'$ and $X_t \to X'$, applying the mean value theorem to the left hand side of \eqref{eq:armijo2}, we have that
\begin{flalign*}
-\langle L([X_t + \alpha_t' d_t]_+C) - B, \mathbb{D}([X_t + \alpha_t' d_t]_+)[d_t] \rangle &=(\alpha_t \beta^{-1})^{-1}(F(X_t)) - F([X_t + \alpha_t \beta^{-1}d_t]_+) \\
&< -\sigma \langle \mathcal{P}(X_t), d_t\rangle
\end{flalign*}
for some $\alpha_t' \in [0, \alpha_t\beta^{-1}]$. Taking the limit as $k\to \infty$, we have that $\alpha_t'\to 0$ and $\mathbb{D}([X_t + \alpha_t' d_t]_+)[d_t] \to \mathcal{P}(X')$, which implies 
$$
-\langle \mathcal{P}(X'), d' \rangle \leq -\sigma \langle \mathcal{P}(X'), d' \rangle, \:\: \text{i.e. } -(1-\sigma)\langle \mathcal{P}(X'), d' \rangle \leq 0
$$
Since $\sigma < 1$, it follows that 
\begin{equation}\label{eq:gd_convergence}
-\langle \mathcal{P}(X'), d' \rangle = ||\mathcal{P}(X')||_F^2 \leq 0
\end{equation}
which contradicts the non-stationarity of $X'$. Hence, the limit point $\hat{X}$ is a stationary point. \qedsymbol

\vskip 0.2in
\bibliography{main}

\end{document}